\documentclass[letterpaper]{article} 
\usepackage{aaai25}  
\usepackage{times}  
\usepackage{helvet}  
\usepackage{courier}  
\usepackage[hyphens]{url}  
\usepackage{graphicx} 
\def\UrlFont{\rm}  
\usepackage{natbib}  
\usepackage{caption} 
\usepackage{algorithm}
\usepackage{algorithmic}
\usepackage{colortbl}
\usepackage{xcolor}
\usepackage{amssymb}
\usepackage{makecell}

\usepackage{newfloat}
\usepackage{listings}
\DeclareCaptionStyle{ruled}{labelfont=normalfont,labelsep=colon,strut=off} 
\floatstyle{ruled}
\newfloat{listing}{tb}{lst}{}
\floatname{listing}{Listing}
\title{\LARGE Diffusion Synthesis: Data Factory with Minimal Human Effort Using VLMs}

\author{\centering \hspace{1.5cm}  Jiaojiao Ye$^{\star}$, 
    Jiaxing Zhong$^{\star}$, 
    Qian Xie$^{\dagger}$, 
    Yuzhou Zhou$^{\star}$, 
    Niki Trigoni$^{\star}$, 
    Andrew Markham$^{\star}$}

\affiliations{$^{\star}$University of Oxford \\
      $^{\dagger}$University of Leeds}

\begin{document}

\maketitle

\begin{abstract}
Generating enough and diverse data through augmentation offers an efficient solution to the time-consuming and labor-intensive process of collecting and annotating pixel-wise images.
Traditional data augmentation techniques often face challenges in manipulating high-level semantic attributes, such as materials and textures. In contrast, diffusion models offer a robust alternative, by effectively utilizing text-to-image or image-to-image transformation. However, existing diffusion-based methods are either computationally expensive or compromise on performance. To address this issue, 
we introduce a novel training-free pipeline that integrates pre-trained ControlNet and Vision-Language Models (VLMs) to generate synthetic images paired with pixel-level labels. This approach eliminates the need for manual annotations and significantly improves downstream tasks. To improve the fidelity and diversity, we add a Multi-way Prompt Generator, Mask Generator and High-quality Image Selection module. Our results on PASCAL-$5^i$ and COCO-$20^i$ present promising performance and outperform concurrent work for one-shot semantic segmentation.
\end{abstract}

\section{Introduction}
\label{sec:intro}
The rapid advancements in machine learning and computer vision have significantly improved the performance of semantic segmentation task. However, the success of these models remains heavily dependent on large-scale, manually annotated datasets, which are often time-consuming, expensive and labor-intensive to produce. For instance, labeling a single urban image in the Cityscapes dataset~\cite{cordts2016cityscapes} can take up to 60 minutes, highlighting the complexity of this task. Furthermore, in data-scarce or imbalanced domains, such as medical imaging ~\cite{hashimoto2020multi, Willemink2020PreparingMI} or specific industrial applications \cite{article}, acquiring sufficient labeled data becomes an even greater challenge.

To address this issue, previous research has extensively 
explored standard model-free data augmentation ~\cite{zhang2017mixup,inoue2018data,yun2019cutmix,bochkovskiy2020yolov4,dwibedi2017cut}. However, basic transformation fails to produce high-level semantic attributes, such as textures, animal species, or changes in perspective. In contrast, recent studies have focused on leveraging generative models for diverse, large-scale synthetic data generation as a means to augment datasets and enhance model performance. Generative Adversarial Networks (GANs) and Diffusion Models (DMs) have been prominently used to create synthetic data that mimics real-world distributions, thereby enriching the training datasets without the extensive need for manual annotation 
~\cite{nguyen2023dataset, feng2023diverse, trabucco2023effective, wu2023diffumask}.  
For instance, DA-Fusion~\cite{trabucco2023effective} provides a way to address the lack of diversity in the context of high-level semantic attributes, by using image-to-image transformations. 
Similarly, Dataset Diffusion~\cite{nguyen2023dataset} generates pixel-level semantic segmentation labels using text-to-image Stable Diffusion (SD) ~\cite{rombach2021highresolution}, cross-attention, and self-attention mechanisms, eliminating the need for labor-intensive pixel-wise annotation. 
However, these current methods either require extensive training or fail to generate data for complex scenes. For instance, in Dataset Diffusion, when the text prompt involves three or more objects, the diffusion model may generate images with fewer objects, resulting in image content that does not fully match the prompt. DA-Fusion requires performing textual inversion for novel concepts for different domain, which is computationally expensive and time-consuming. 

\begin{figure}[ht]
\centering\includegraphics[width=0.45\textwidth]{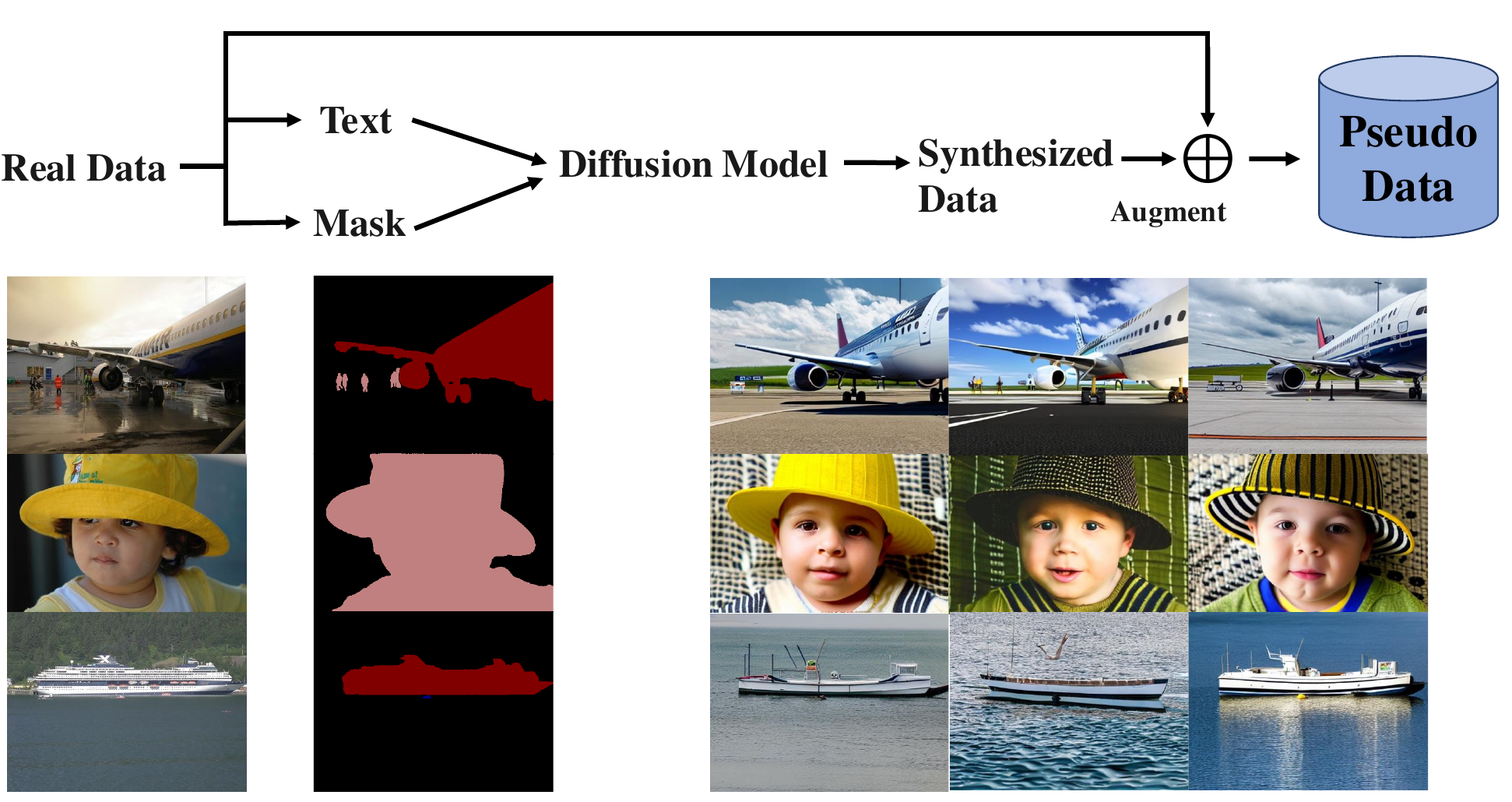} 
\caption[How our Diffusion-Synthesis framework.]{Given an image dataset, our pipeline can generate K synthetic images paired with pixel-level labels using VLMs and a pre-trained Controlnet checkpoint. The resulting synthetic images are mixed with real images for training downstream tasks.}
\label{fig:pipeline}
\end{figure}

In this work, we propose a training-free data augmentation pipeline, \textbf{Diffusion Synthesis}, for efficient and high-quality data generation, as illustrated in Fig. \ref{fig:pipeline}. Our approach leverages pre-trained Vision-Language Models (VLMs) and publicly available off-the-shelf conditional diffusion models to synthesize diverse, high-fidelity data with corresponding semantic labels, minimizing human effort and cost. Compared to previous methods, our approach effectively handles complex cases involving multiple objects, aided by mask guidance and detailed prompts. Additionally, we introduce a Multi-way Prompt Generator module, a Mask Generator, and a High-quality Image Selection module to ensure that the generated data is semantically and visually diverse, as well as physically flexible and realistic. 
Finally, we test our method on few-shot semantic segmentation (FSS), a challenging task that involves predicting dense masks for new classes with only a limited number of annotations. Our method achieves 69.1\% mIoU on PASCAL-$5^i$ and 43.4\% mIoU on COCO-$20^i$ respectively, surpassing the prior work Dataset Diffusion~\cite{nguyen2023dataset} in both perceptual quality and downstream performance.




\begin{figure*}[ht!]
\centering\includegraphics[width=0.7\textwidth]{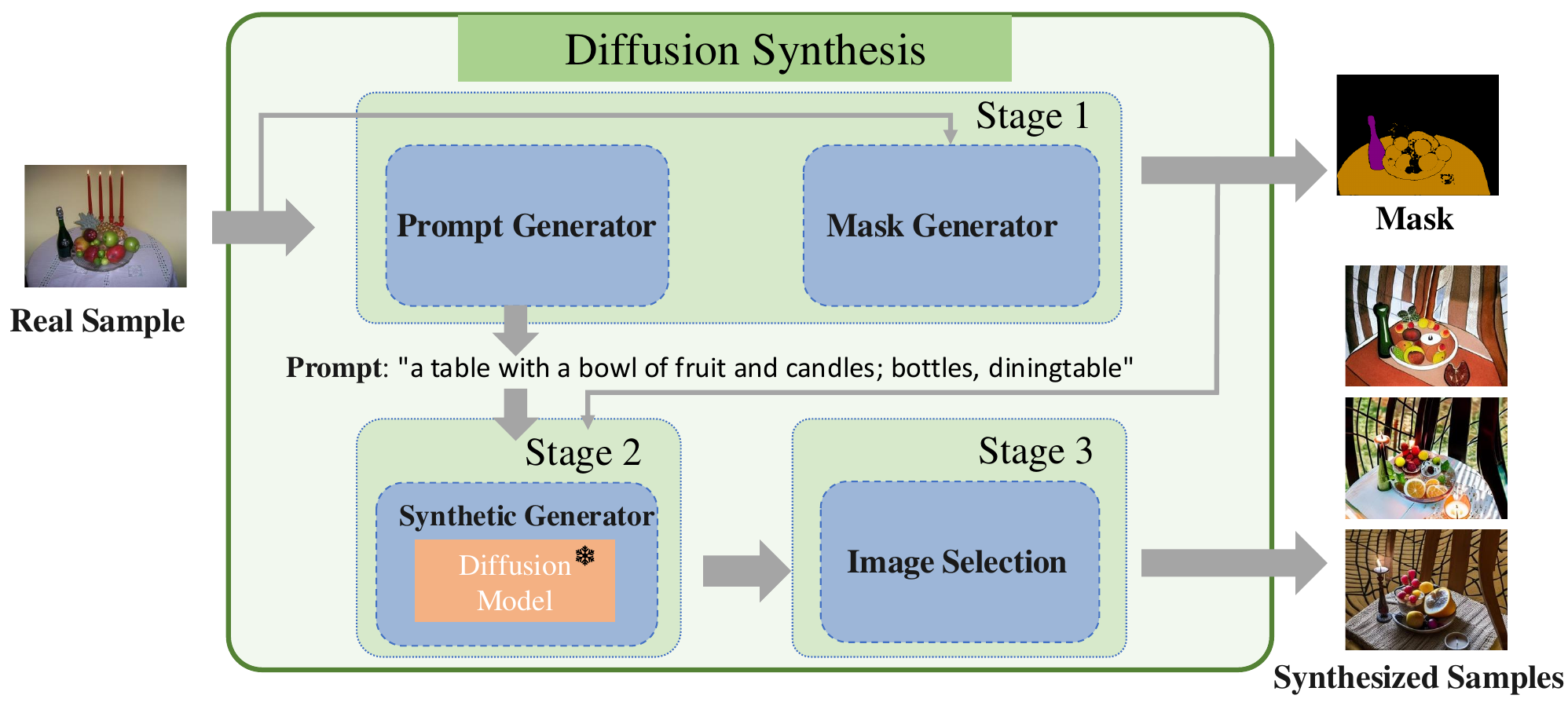} 
\caption[Diffusion-Synthesis module framework.]{Overview of the Diffusion Synthesis framework, consisting of three seperated stages. Given an image dataset, our pipeline can generate $K$ synthetic images paired with pixel-level masks through image-to-image translation using VLMs and a pre-trained Diffusion Model. Resulted synthetic images are combined with real images for training downstream segmentation tasks.}
\label{fig:framework}
\end{figure*}

\section{Related Work}
\label{sec:related}


\subsection{Data Augmentation} 
\label{ssec:data augmentation} 
Data augmentation in computer vision tasks usually involves systematically transforming existing images to create diverse training datasets, thereby enhancing the robustness and performance of vision models. Based on their underlying mechanisms, image augmentation methods can be broadly categorized into model-free and model-based approaches.

\textbf{Model-Free image augmentation.} Model-free approaches do not rely on pre-trained models but instead utilize simple image processing techniques to generate new training images.
These methods can be further divided into non-instance-level~\cite{zhang2017mixup,inoue2018data,yun2019cutmix,bochkovskiy2020yolov4,dabouei2021supermix} and instance-level~\cite{georgakis2017synthesizing,dwibedi2017cut,dvornik2018modeling,ghiasi2021simple} augmentations.
Among them, Mixup~\cite{zhang2017mixup} generates augmented images by taking linear interpolations of pairs of examples and their labels, encouraging the model to behave linearly in-between training examples.
Following Mixup, CutMix~\cite{yun2019cutmix} replaces a portion of an image with a patch from another image while adjusting the labels accordingly, achieving significant improvements in both image classification and localization tasks.
Based on the instance-level mask, Dwibedi et al.~\cite{dwibedi2017cut} present a method where object instances are cut from images and pasted onto different backgrounds. This simple yet effective technique helps in synthesizing new training examples for object detection tasks.
In a similar way, Ghiasi et al.~\cite{ghiasi2021simple} enhance instance segmentation performance by randomly pasting object instances into other images, effectively increasing the diversity of the training data without sophisticated operations.

\textbf{Model-based image augmentation.}
In contrast to model-free methods, model-based image augmentation techniques~\cite{douzas2018effective, ali2019mfc,li2020shape, hong2021stylemix, zheng2021generative,xu2022style} leverage generative models to create new samples, typically involving more complex computations but potentially generating more realistic and diverse images.
For instance, Tran et al.~\cite{tran2017bayesian} propose a Bayesian framework that integrates data augmentation directly into the model training process, allowing for uncertainty quantification and robust learning even with limited data.
Inspired by Mixup, Stylemix~\cite{hong2021stylemix} separates the content and style of images to create new samples.

\subsection{Diffusion Model-based Image Editing}
\label{ssec:DMIE} 
Distinct from image generation, which synthesizes new images from minimal inputs, image editing entails modifying the appearance, structure, or content of an existing image.
Recently, the introduction of diffusion models~\cite{ho2020denoising, song2020denoising,  nichol2021improved} has led to a pivotal advancement in the field of image editing.
Depending on the learning strategy, diffusion model-based image editing methods can be divided into three categories: training-based approaches~\cite{wang2023stylediffusion,xu2024cyclenet,huang2024smartedit}, testing-time fine-tuning approaches~\cite{choi2023custom, mokady2023null,dong2023prompt,zhang2023sine}, and training and fine-tuning free approaches~\cite{couairon2022diffedit,pan2023effective}.
Readers interested in this topic are encouraged to refer to the most recent related review paper~\cite{huang2024diffusion}.

\subsection{Diffusion Model-based Augmentation}
\label{ssec:DMDA} 
Training models with synthetic images \cite{li2022bigdatasetgan, nguyen2023dataset, zhang2021datasetgan, feng2023diverse, trabucco2023effective, mandi2022cacti} are rapidly gaining popularity and undergoing rapid development. Existing research advocates leveraging generative AI, such as GANs \cite{li2022bigdatasetgan, zhang2021datasetgan}, text-to-image diffusion models \cite{nguyen2023dataset, feng2023diverse, liu2023matcher}, for data augmentation. For instance, DA-Fusion \cite{trabucco2023effective} generates diverse images by altering their semantics using off-the-shelf diffusion models and generalizes to novel visual concepts from a few labelled examples, yielding significant improvements in few-shot image classification tasks. Conversely, Dataset Diffusion \cite{nguyen2023dataset}, on the other hand, introduces a technology for pseudo-label generation and employs more complex text prompts, facilitating the creation of intricate and realistic images.

However, these approaches either focus on image-level category labels or struggle in complex scenarios, such as when multiple objects are present in a single image or objects are closely intertwined. Moreover, existing methods for dataset synthesis or augmentation typically require fine-tuning or extensive training, which demands significant computational resources and time. In contrast, we propose a training-free dataset synthesis pipeline that automatically generates realistic images and pixel-level pseudo-semantic segmentation labels, eliminating the need for manual labeling or prompt engineering. This approach streamlines the data preparation process, significantly reducing human effort while maintaining high-quality, detailed annotations.

\section{Methodology}
\label{sec:method}

\subsection{Problem Setting}
The goal of this work is to generate a dataset, denoted as $\mathcal{D} 
= \{\mathcal{I}_{i}^{'}, \mathcal{M}_{i}, \mathcal{P}_{i} \}_{i=1}^{N}$, which consists of high-fidelity images $\mathcal{I}_{i}^{'}$, pixel-level semantic labels $\mathcal{M}_{i}$, and corresponding captions $\mathcal{P}_{i}$, based on a set of real image $ \{\mathcal{I}_{i}\} $. The purpose of constructing this dataset is to provide physically accurate and visually diverse information to train a few-shot semantic segmentation without human annotation.

\subsection{Overall Structure}
\label{ssec:pipeline}

Our Dataset Synthesis is a pipeline for generating diverse and high-quality images and masks without training, consisting of three components: A Multi-way Prompt Generator (MPG), a Mask Generator (MG) and a High-quality Image Selection (HIS) module. As illustrated in Fig. \ref{fig:framework}, our Dataset Synthesis consists of three stages. In the first stage, given a set of images $ \{ \mathcal{I}_{i} \in \mathcal{R}^ {H \times W \times 3} \}_{i=1}^{N} $, we extract descriptive prompts $\mathcal{P}$ using a MPG. The Visual Language Model, BLIP \cite{li2022blip} is firstly employed to generate captions for source images. These captions are then either directly used as prompts or augmented with object class labels to ensure accurate object generation. Real captions, such as those from COCO \cite{lin2014microsoft}), can also be utilized to avoid missing objects in the prompt generation. Subsequently, our MG, primarily composed of a zero-shot open-vocabulary object detector Dino \cite{zhang2023dino} and box-guided semantic segmenter SAM \cite{kirillov2023segment}, predicts masks $\mathcal{M}_{i}$, which serves as pseudo-labels for the final dataset. In the second stage, a frozen ControlNet \cite{zhang2023adding} is used to generate $K$ augmented versions of each image, conditioned on the previously generated semantic masks and prompts. These synthetic images maintain the same pixel-level classes but vary in semantic contexts, such as colors, lighting and backgrounds. Finally, HIS module filters artifacts to form the final augmented dataset $\mathcal{D}$. These synthetic images and their corresponding semantic annotations are then mixed with real data to train downstream semantic segmentation models.

\subsection{Multi-way Prompts Generator}
\label{ssec:syn}

\begin{figure}[ht]
    \centering
    \begin{minipage}[t]{0.23\textwidth}
        \includegraphics[width=0.3\textwidth]{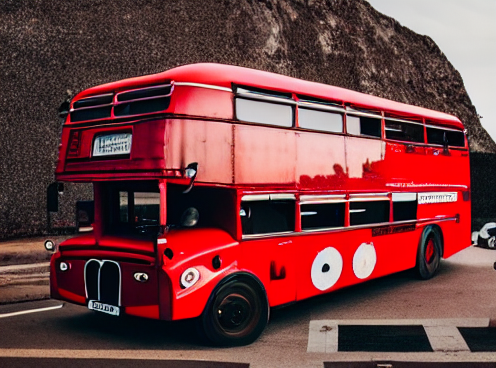}
        \includegraphics[width=0.3\textwidth]{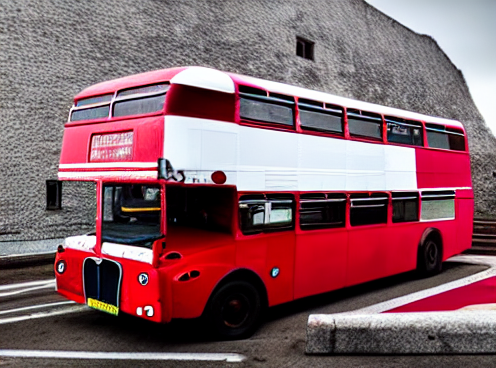}
        \includegraphics[width=0.3\textwidth]{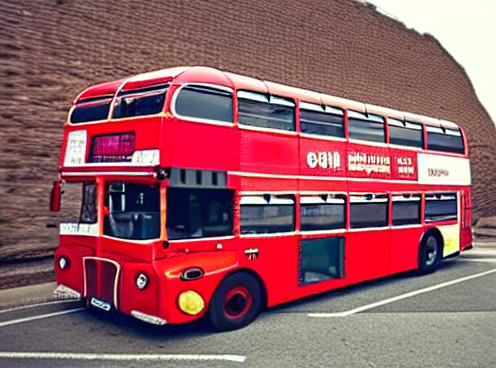}
    \captionsetup{labelformat=empty} 
    \caption*{Prompt: \textit{a red bus parked next to a white bus; bus}.}
    \end{minipage}
    \hfill
    \vrule width 1pt 
    \hfill
    \begin{minipage}[t]{0.23\textwidth}
        \includegraphics[width=0.3\textwidth]{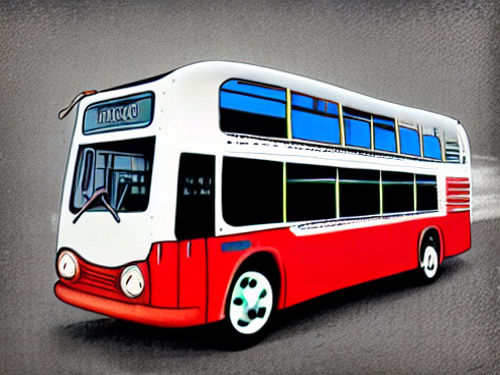}
        \includegraphics[width=0.3\textwidth]{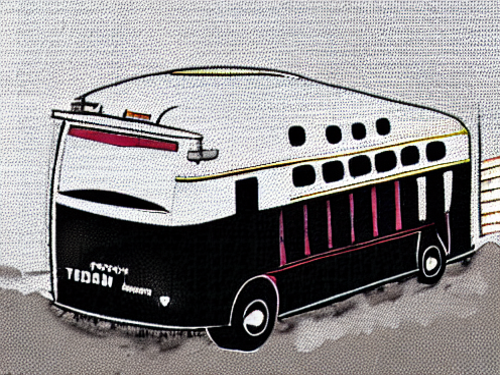}
        \includegraphics[width=0.3\textwidth]{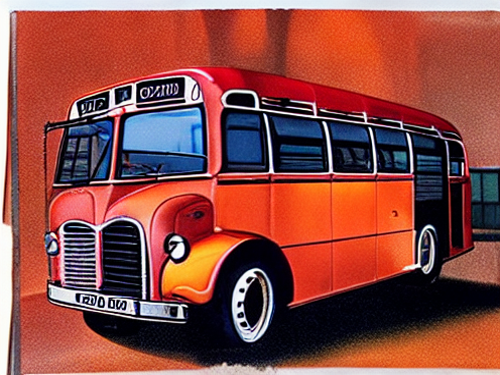}
    \captionsetup{labelformat=empty} 
    \caption*{Prompt: \textit{An image of bus}.}
    \end{minipage}

    \begin{minipage}[t]{0.23\textwidth}
        \includegraphics[width=0.3\textwidth]{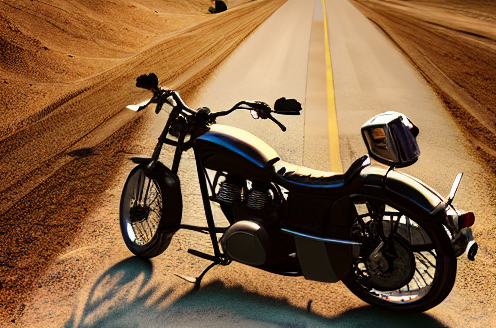}
        \includegraphics[width=0.3\textwidth]{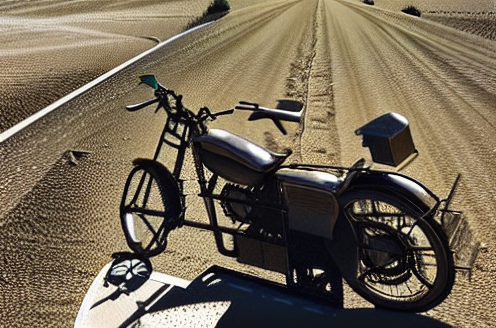}
        \includegraphics[width=0.3\textwidth]{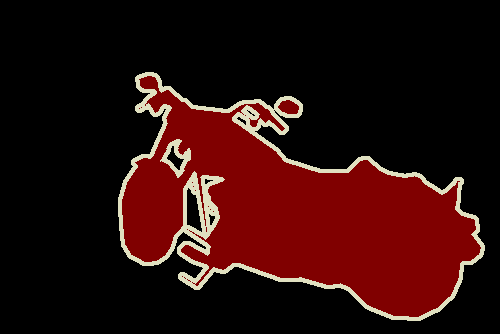}
    \captionsetup{labelformat=empty} 
    \caption*{Prompt: \textit{A motorcycle parked on a dirt road; motorbike}.}
    \end{minipage}
    \hfill
    \vrule width 1pt 
    \hfill
    \begin{minipage}[t]{0.23\textwidth}
        \includegraphics[width=0.3\textwidth]{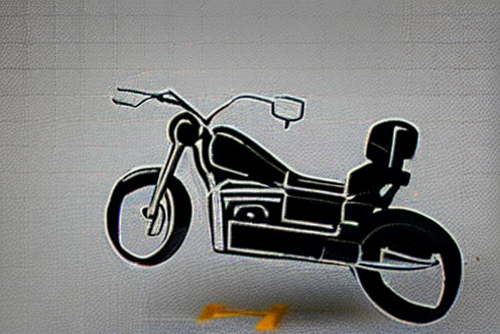}
        \includegraphics[width=0.3\textwidth]{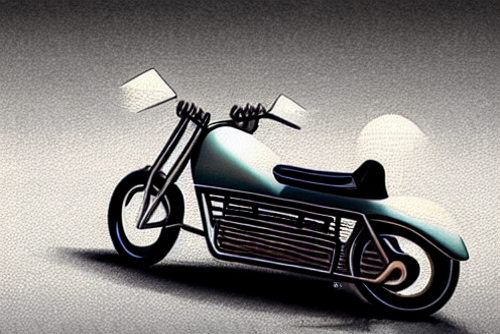}
        \includegraphics[width=0.3\textwidth]{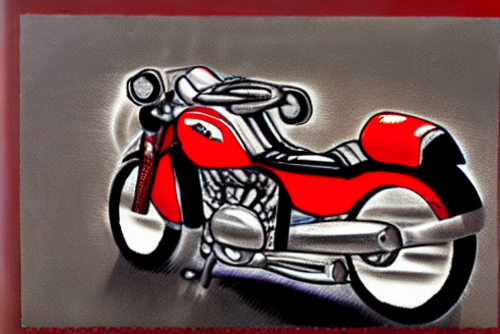}
    \captionsetup{labelformat=empty} 
    \caption*{Prompt: \textit{An image of motorbike}.}
    \end{minipage}
    
    \begin{minipage}[t]{0.23\textwidth}
        \includegraphics[width=0.3\textwidth]{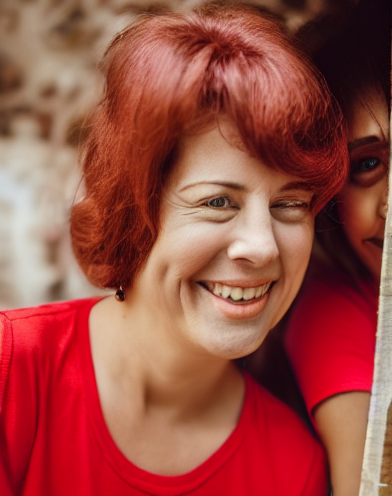}
        \includegraphics[width=0.3\textwidth]{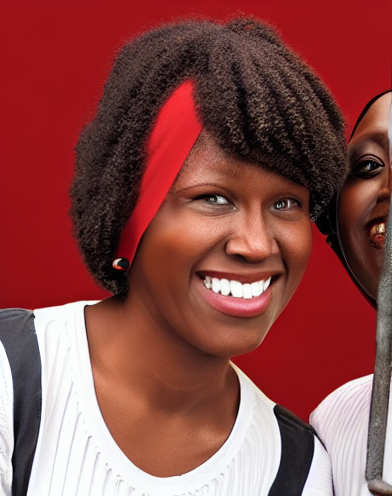}
        \includegraphics[width=0.3\textwidth]{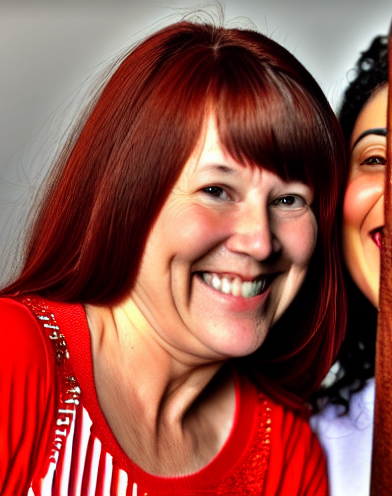}
    \captionsetup{labelformat=empty} 
    \caption*{Prompt: \textit{Two women in red shirts posing for a photo; person}.}
    \end{minipage}
    \hfill
    \vrule width 1pt 
    \hfill
    \begin{minipage}[t]{0.23\textwidth}
        \includegraphics[width=0.3\textwidth]{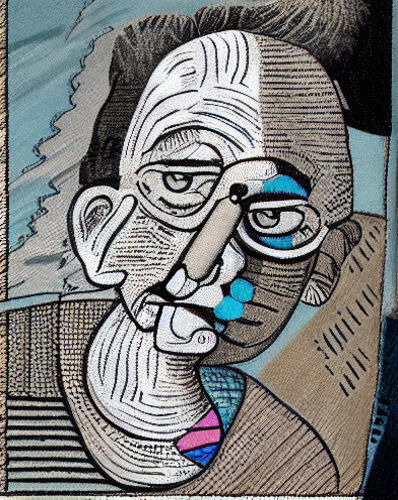}
        \includegraphics[width=0.3\textwidth]{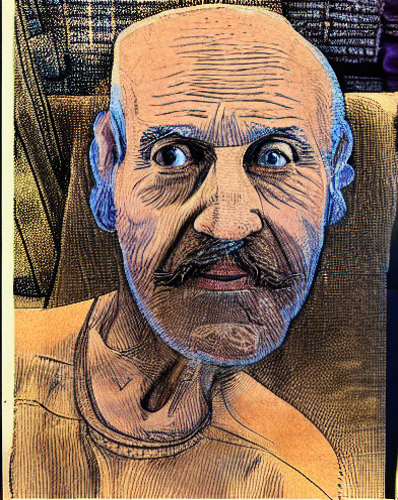}
        \includegraphics[width=0.3\textwidth]{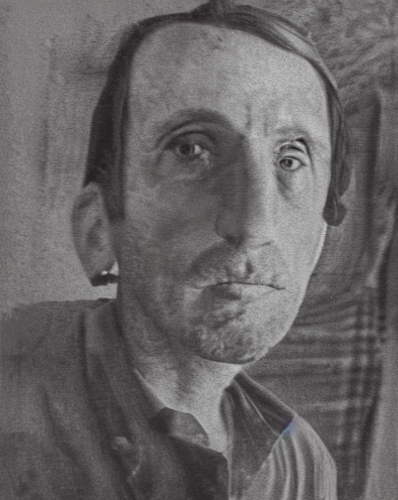}
    \captionsetup{labelformat=empty} 
    \caption*{Prompt: \textit{An image of person}.}
    \end{minipage}

    \begin{minipage}[t]{0.23\textwidth}
        \includegraphics[width=0.3\textwidth]{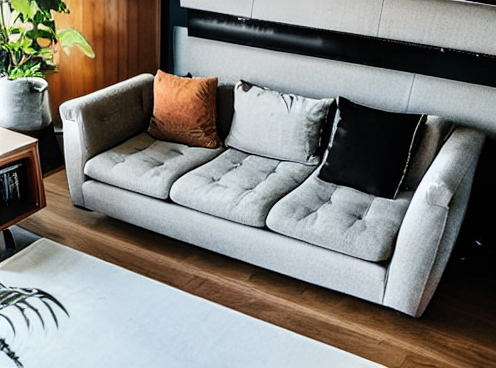}
        \includegraphics[width=0.3\textwidth]{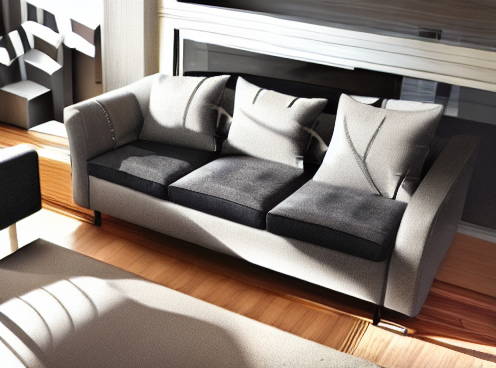}
        \includegraphics[width=0.3\textwidth]{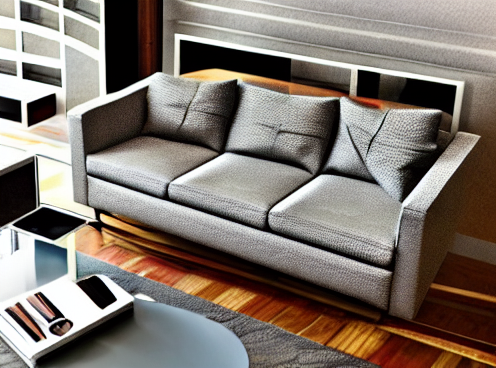}
    \captionsetup{labelformat=empty} 
    \caption*{Prompt: \textit{A living room with a couch and a coffee table; pottedplant, sofa, chair}.}
    \end{minipage}
    \hfill
    \vrule width 1pt 
    \hfill
    \begin{minipage}[t]{0.23\textwidth}
        \includegraphics[width=0.3\textwidth]{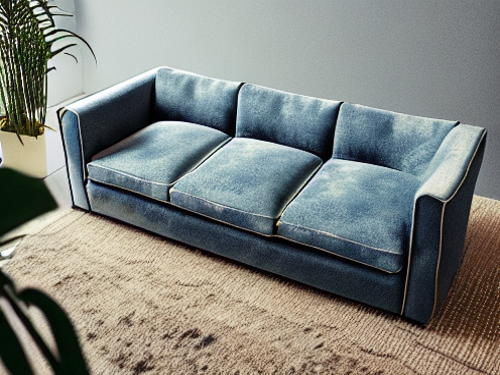}
        \includegraphics[width=0.3\textwidth]{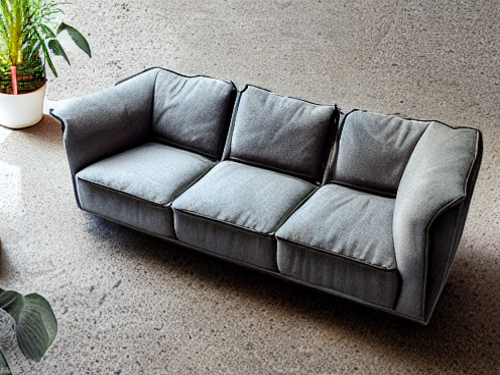}
        \includegraphics[width=0.3\textwidth]{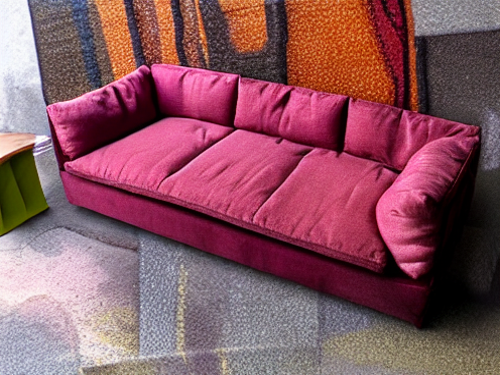}
    \captionsetup{labelformat=empty} 
    \caption*{Prompt: \textit{An image of pottedplant, sofa, chair}.}
    \end{minipage}

    \caption{Ablation study of Prompt. Visualization of the diverse and informative diffusion-based augmented images. Columns one to three show images produced using our prompt generator, while columns four to six present images generated using a simple class method.}
    \label{fig:promptablation}
\end{figure}

There are several compelling reasons for introducing the Multi-Way Prompts Generator (MPG). First, traditional prompt representations in the text-to-image generation, i.e., ``an image of [class name]," are overly simplistic and fail to encode detailed information. For instance, they omit spatial relationships of objects and contextual descriptions, such as backgrounds (dirty roads) illustrated in the second line of Fig. \ref{fig:promptablation}. Second, image captions are not guaranteed to capture the desired object class, and the terminology used in captions may not align with the intended class defined in the dataset, like "bicycles" instead of ``bikes" or ``table" instead of ``diningtable". This inadequacy results in an insufficient number of text prompts and mismatching for certain classes, negatively impacting the generation process for those specific classes.

Given a set of images, the Multi-Way Prompts Generator offers multiple methods to generate descriptive and accurate prompts. Initially, we utilize the state-of-the-art image captioning model BLIP \cite{li2022blip} to generate captions for each image, which serve as the base Prompts $\mathcal{P}$. Additionally, we extend these captions by appending class labels, creating new text prompts that explicitly incorporate all target classes, $\mathcal{C} = [ \mathcal{C}_1, \mathcal{C}_2, \cdot, \mathcal{C}_n ]$, where $\it n$ represents the number of classes associated with the image. This is achieved through a text appending operation, or class-prompt appending technique, represented as $ \mathcal{P}^{'}_i = \{ \mathcal{P}_i; \mathcal{C}_i \} $. For instance, in the last case of the left image in Fig. \ref{fig:promptablation}, the final text prompt would be ``A living room with a couch and a coffee table; pottedplant, sofa, chair". 

Simultaneously, we observed that generated images exhibit realistic and diverse semantics, even when using the same prompt and mask. For instance, the object ``sofa" in the final row of Fig. \ref{fig:promptablation} maintains the same contour but varies in materials, colors, and environments. Additionally, for more complex subjects like a person, our prompt engineering technique ensures the generation of versatile synthetic images and successfully captures detailed features such as teeth and eyes. In contrast, the text-to-image generator struggles to produce realistic images when guided only by a simple class prompt.

\begin{figure}[ht]
\centering\includegraphics[width=0.45\textwidth]{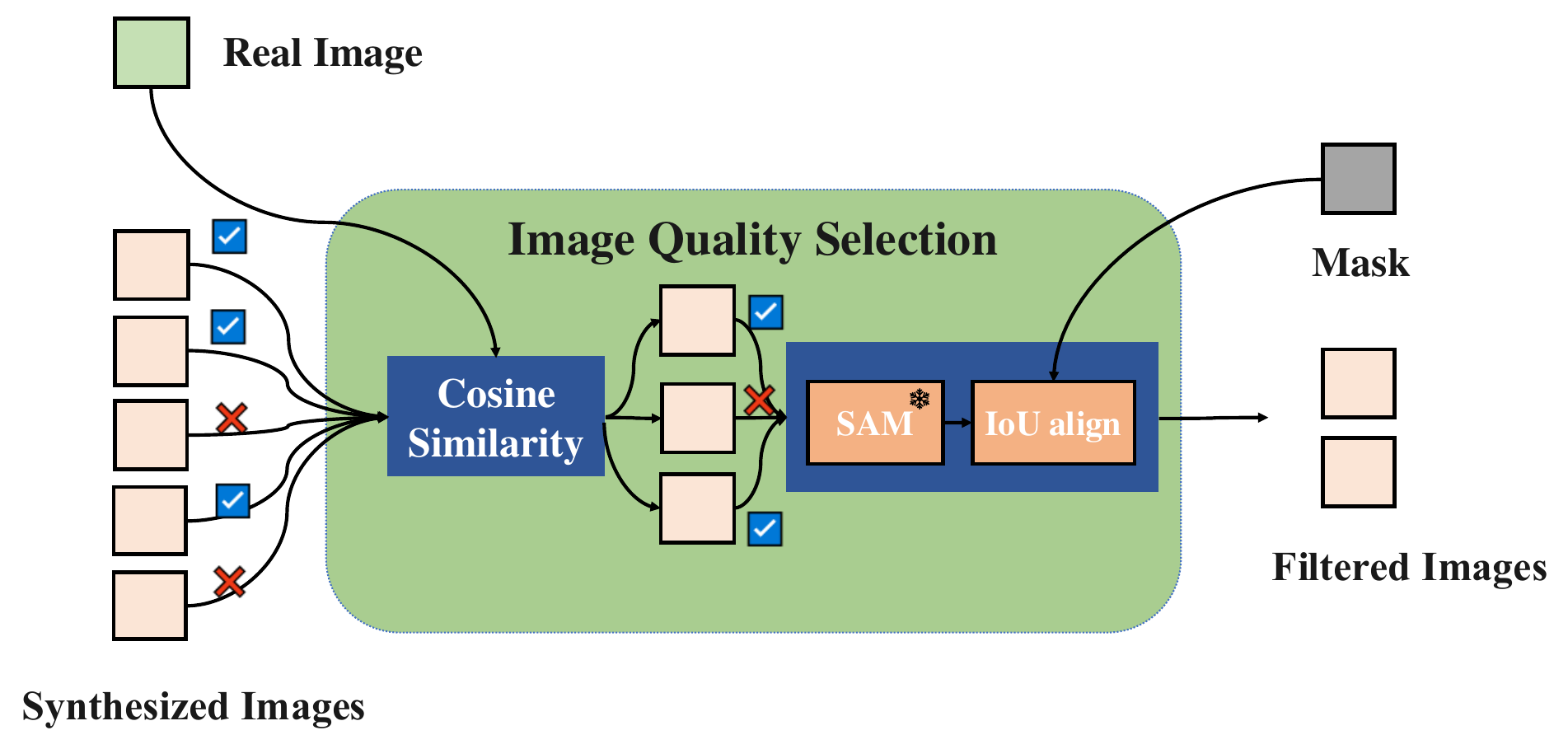} 
\caption[Diffusion-Synthesis module framework.]{Illustration of our selection module. This module has two sub-processes: cosine similarity filtration and foundation-model driven matching. This module takes the source image, synthesized image, and previous mask as input, then produces high-quality filtered images as output.}
\label{fig:selection}
\end{figure}

\subsection{Conditional Generator via Diffusion}
\label{ssec:gen}
We develop our dataset generator using ControlNet \cite{zhang2023adding} by leveraging its self-attention and cross-attention layers. 
Specifically, given a text prompt $ \mathcal{P^{'}}$ and a condition mask $\mathcal{M}$, the module first encodes the text through a text encoder, generating a text embedding $ e \in \mathbb{R}^{\Lambda} \times d_{e} $, where \( \Lambda \) represents the text length and \( d_{e} \) denotes the embedding dimension. The module then concatenates these features, producing the final latent state \( z_{0} \) after $T$ denoising steps. Finally, the latent state
$z_0$ is decoded by an image decoder, resulting in a high-quality synthetic image $\mathcal{I}^{'}$ that corresponds to the description but with diverse semantics interpretations. Thus, the augmented image can then be generated as described in Equation \ref{eq:dm}.
Here, $ n \sim \mathcal{N}(0, 1)$ denotes the sampled noise. 

\begin{equation}
  \mathcal{D} ({\mathcal{I}} _i) =  \mathcal{G}(\mathcal{I}_i, \mathcal{P}'_i, n \mid \mathcal{M} ) 
\label{eq:dm}
\end{equation}

\subsection{High-quality Image Selection}
\label{ssec:his}
To balance the diversity and quality of synthetic data used for training, we implement a human-free image selection strategy to identify and retain high-quality generated samples. As illustrated in Fig. \ref{fig:selection}, our High-quality Image Selection (HIS) module consists of two sub-components: cosine similarity filtration and foundation model-driven matching. 

First, to retain high fidelity in the synthetic data generation process, we introduce a cosine similarity-based selection approach. Specifically, we compute the cosine similarity between the real sample $\mathcal{I}_i$ and each augmented image ${\mathcal{I}^{'}}_i$. Generated images that exhibit a cosine similarity greater than a predefined threshold $\epsilon$ are retained, as stated in Equation \ref{eq:cos}. This method ensures that the selected synthetic images closely resemble the original data, thereby enhancing the overall quality and relevance of the augmented dataset.

\begin{equation}
  {\mathcal{I}'} _i =  cos(\mathcal{I}_i, \mathcal{D} ({\mathcal{I}} _i) )> \epsilon 
\label{eq:cos}
\end{equation}


Additionally, we employ a foundation model-driven matching block to ensure alignment between masks and generated images. Mismatches can occur because the generator sometimes fails to preserve fine-grained features, such as the presence of multiple objects, as noted in Dataset Diffusion \cite{nguyen2023dataset}. Specifically, we use a segmenter to mask the synthesized image and obtain a mask prediction $\mathcal{M}'$. We then compare the alignment between the predicted mask $\mathcal{M}'$ and the original pseudo mask $\mathcal{M}$. Synthetic images with low pixel-level label accuracy (i.e., mean Intersection over Union, mIoU) are discarded if there is a significant discrepancy between the predicted and the original masks. This process further improves the quality of the synthetic dataset $\mathcal{D}'$.

\subsection{Controlling Balance between Real and Synthetic Data} 
\label{ssec:ratio}
Training models on synthetic images often risks overemphasizing spurious attributes and biases due to imperfections inherent in generative models. A common approach to mitigate this issue involves assigning different sampling probabilities to real and synthetic images, addressing the imbalance between these data types, as proposed in \cite{he2022synthetic} \cite{trabucco2023effective}. In our work, we adopt a similar strategy to balance real and synthetic data, as detailed in Equation \ref{eq:ratio}, where $\alpha$ represents the probability that a synthetic image occupies the $l-th$ position in the mini-batch of images B.

\begin{equation}
P(\mathcal{I}^{'}_{i,j} \in B_{l}) = \alpha
\label{eq:ratio}
\end{equation}

Let $\mathcal{I}^{'}_{i,j} \in \mathcal{R}^ {H \times W \times 3}$ represents a synthetic image, where $i \in \mathcal{Z}$ specifies the index of a particular image. For each image, we generate $K$ augmentations, resulting in a synthetic dataset $\{\mathcal{I}^{'} \} \in \mathcal{R}^ {N \times K \times H \times W \times 3}$. Indices $i$ and $j$ are sampled uniformly from the $N$ real images and their $K$ augmented versions, respectively. Given indices $i$ and $j$, a real image $\mathcal{I}$ is added to the batch B with probability $(1 - \alpha)$; otherwise, its augmented version $\mathcal{I}'$ is added. 
Thus, we can obtain a large number of augmented samples with richer visual appearance variation while preserving key semantics.

\section{Experiments}
\label{sec:experiment}

\begin{table*}[ht]
\centering
\setlength{\tabcolsep}{1mm}
    \begin{tabular}{lll|lllll}
    \hline 
    \multicolumn{3}{l|}{} & \multicolumn{5}{c}{1-shot (mIoU) } \\
    \textbf{Method} & \textbf{Conference}& \textbf{Training-free} & Fold-0 & Fold-1 & Fold-2 & Fold-3 & Mean$\uparrow$
    \\ \hline
    SSP \cite{fan2022ssp} & ECCV’22 & & 60.5 & 67.8 & 66.4 & 51.0 & 61.4 
    \\
    MIANet \cite{yang2023mianet} & CVPR’23 & & 68.5 & 75.8 & 67.5 & 63.2 & 68.7 
    \\ 
     SCCAN \cite{xu2023self} & ICCV’23 &  & 68.3 & 72.5 & 66.8 & 59.8 & 66.8 
    \\ 
     DCAMA \cite{shi2022dense} & ECCV'22 & & 67.5 & 72.3 & 59.6 & 59.0 & 64.6 
    \\ 
    BAM\textsuperscript{$\dagger$} \cite{lang2022bam} & CVPR'22 &  & 65.7 & 71.4 & 65.6 & 58.9 & 65.4 
     \\ 
    synth-VOC\textsuperscript{$\dagger$} \cite{nguyen2023dataset} & NeurIPS'23 & & 66.2(+0.5) & 74.4\textbf{(+3.0)} & 64.0(-1.6) & 59.0(+0.1)  & 65.9(+0.5)
    \\
    \cline{1-8}
    \rowcolor{lightgray} 
    Ours (+ BAM) & & \checkmark  &  68.8\textbf{(+3.1)} & 73.3(+1.9) & 66.7(+1.1) & 61.7\textbf{(+2.8)} & 67.6(+2.2)
    \\
    \cline{1-8}
    HDMNet\textsuperscript{$\dagger$} \cite{peng2023hierarchical} & CVPR'23 &  & 67.1 & 74.9 & 64.5 & 60.7 & 66.8 
    \\
    synth-VOC\textsuperscript{$\dagger$} \cite{nguyen2023dataset} & NeurIPS'23 & & 69.3(+2.2) & 75.9(+1.0) & 64.8(+0.3) & 60.8(+0.1) & 67.7(+0.9)
    \\
    \cline{1-8}
    \rowcolor{lightgray} 
    Ours (+ HDMNet) & & \checkmark & 70.2\textbf{(+3.1)} & 75.3(+0.4) & 69.6\textbf{(+5.1) }& 61.2(+0.5) & 69.1\textbf{(+2.3)}
    \\
    \cline{1-8}

    \hline
    \end{tabular}
    \caption{Results of 1-shot segmentation learning on PASCAL-$5^i$ using mIoU (\%) evaluation metric before and after applying our method on BAM and HDMNet. Bold indicates the best performance. $\dagger$: Reproduced following the official configuration with 1,000 episodes.}
    \label{table:fsspascal}
\end{table*}

\begin{table*}[ht]
\centering
\setlength{\tabcolsep}{1mm}
    \begin{tabular}{lll|lllll}
    \hline 
    \multicolumn{3}{l|}{} & \multicolumn{5}{c}{1-shot (mIoU)} \\
    \textbf{Method} & \textbf{Conference} & \textbf{Training-free} & Fold-0 & Fold-1 & Fold-2 & Fold-3 & Mean$\uparrow$ 
    \\ \hline
    SSP \cite{fan2022ssp} & ECCV’22 & & 35.5 & 39.6 & 37.9 & 36.7 & 47.4 
    \\
    MIANet \cite{yang2023mianet} & CVPR’23 &  & 42.5 & 53.0 & 47.8 & 47.4 & 47.7 
    \\
    SCCAN \cite{xu2023self} & ICCV’23 &  & 40.0 & 49.7 & 49.6 & 45.6 & 46.3 
    \\ 
    DCAMA \cite{shi2022dense} & ECCV'22 & & 41.9 & 45.1 & 44.4 & 41.7 & 43.3
    \\ 
    BAM\textsuperscript{$\dagger$}  \cite{lang2022bam} & CVPR'22 & & 39.3 & 47.3 & 45.7 & 38.7 & 42.7  
    \\ 
    \cline{1-8}
    \rowcolor{lightgray} 
    Ours (+ BAM) &  & \checkmark & 40.6(+1.3) & 47.4(+0.1) & 45.9(+0.2) & 39.5(+1.1) & 43.4\textbf{(+0.7)}
    \\
    \cline{1-8}
    \hline
\end{tabular}
    \caption{Results of 1-shot segmentation learning on COCO-$20^i$ using mIoU (\%) evaluation metric before and after applying our method on BAM and HDMNet. Bold indicates the best performance. $\dagger$: Reproduced following the official configuration with 10,000 episodes.}
    \label{table:fss}
\end{table*}

\subsection{Datasets and Metrics}
We evaluate our pipeline on two few-shot segmentation benchmark: PASCAL-$5^i$ \cite{shaban2017one} and COCO-$20^i$ \cite{nguyen2019feature}. For each dataset, we perform cross-validation by evenly dividing all classes into four folds. 

\subsubsection{PASCAL-$5^i$.}
PASCAL-$5^{i}$ \cite{shaban2017one} is derived from the PASCAL VOC 2012 \cite{pascal-voc-2012} with additional annotations from SDS \cite{6126343}. It consists of 20 classes.

\subsubsection{COCO-$20^i$.}
COCO-$20^{i}$ \cite{nguyen2019feature} includes 80 classes, based on the MS COCO \cite{lin2014microsoft}. The 80 classes are divided into four folds of 20 classes each, following the same procedure as PASCAL-$5^{i}$ \cite{shaban2017one}. 

\subsubsection{Evaluation metrics.}
We assess the performance of our pipeline using the mean Intersection over Union (mIoU) metric, Fréchet Inception Distance (FID) \cite{heusel2017gans}, and Inception Score (IS) \cite{salimans2016improved}. Lower FID scores indicate higher quality and more diverse generated images, while higher IS values suggest that the generated images are diverse and recognized with high confidence as belonging to specific classes.

\subsubsection{Implementation details.}
Our method is implemented in PyTorch and runs on two Nvidia A10 GPUs with 24 GB of RAM. We use the DINO \cite{zhang2023dino} with a ViT-L/14 \cite{dosovitskiy2020image} as the default box detector and Segment Anything Model (SAM) \cite{kirillov2023segment} with ViT-H as the promptable segmenter of Diffusion Synthesis. We test our augmentation approaches on two state-of-the-art few-shot segmentation models, HDMNet \cite{peng2023hierarchical} and BAM \cite{lang2022bam}. These models are trained in an episode fashion for 50 and 200 epochs on COCO-$20^i$ and PASCAL-$5^i$, respectively, with batch sizes of 4 and 2.

For all few-shot segmentation training, we use AdamW optimizer, as in \cite{zhang2021few}, with a learning rate set to 0.0001. Additionally, weight decay is set to 0.01, and the ``poly" strategy is applied to adjust the learning rate. We use ResNet-50 \cite{he2016deep} as the encoder to extract features with frozen parameters, and PSPNet \cite{zhao2017pyramid} serves as the base learner in all experiments.

\subsection{Quantitative Results}
\label{ssec:quantitative}


We present the segmentation performance of our Diffusion Synthesis on the PASCAL-$5^i$ and COCO-$20^i$ datasets in Table \ref{table:fsspascal} and \ref{table:fss}. Notably, across various challenging domains, there is a consistent and significant improvement in performance for two few-shot segmentation approaches, BAM \cite{lang2022bam} and HDMNet \cite{peng2023hierarchical}. Our synthetic dataset segmentation masks achieve a mIoU of 69.1\% on PASCAL-$5^i$ and 43.4\% on COCO-$20^i$, as shown in table \ref{table:fss}.
These results surpass the previous state-of-the-art achieved by Diffss \cite{tan2023diffss}, improving performance by up to 2.3\%. For instance, BAM's mIoU increased from 65.4\% to 67.6\%, and HDMNet's mIoU rose from 66.8\% to 69.1\%, demonstrating the effectiveness of our augmentation techniques.

\subsection{Qualitative Results}
\label{ssec:qualitative}

\begin{figure}[t!]
\centering\includegraphics[width=0.47\textwidth]{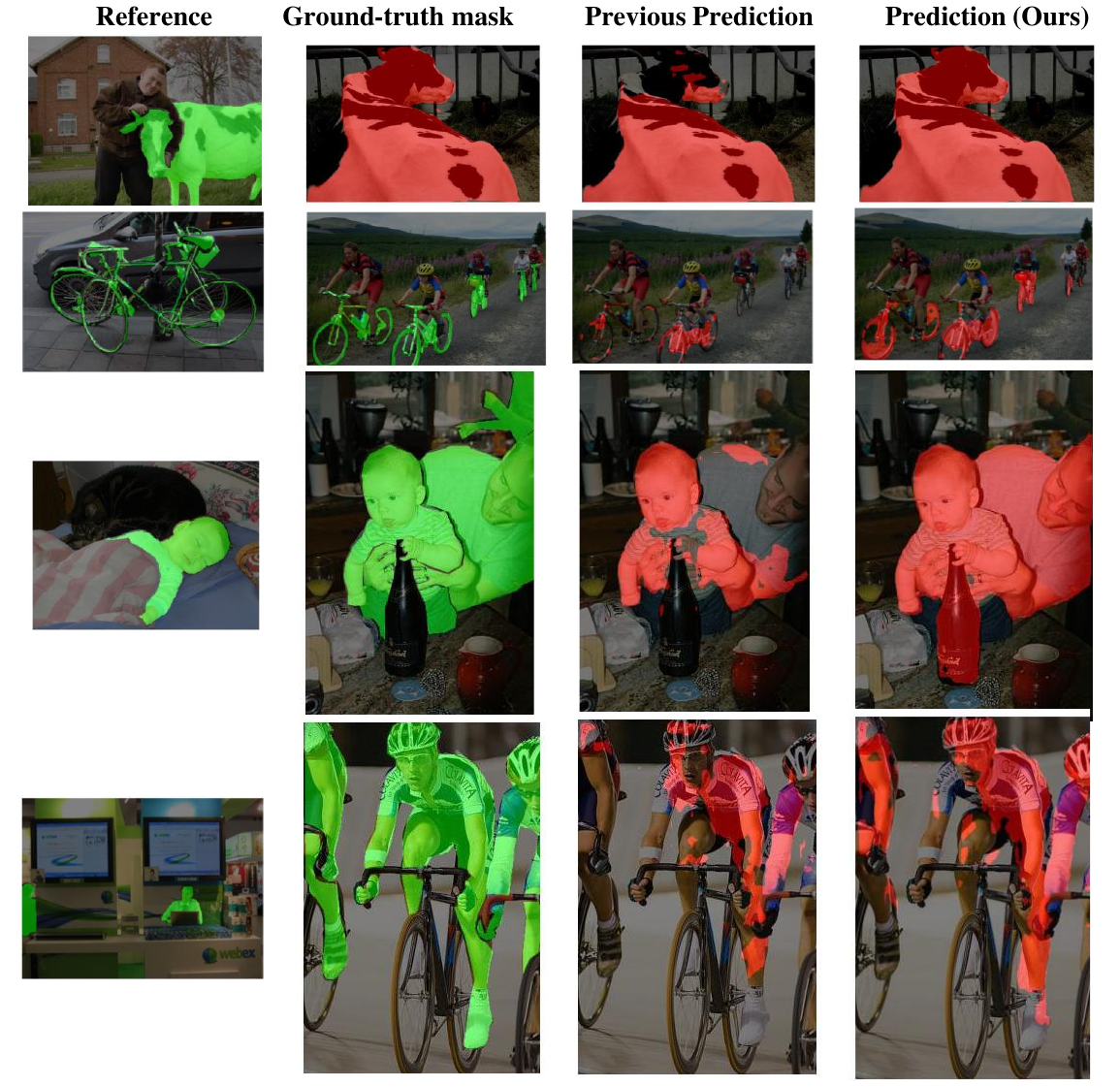}
\centering\includegraphics[width=0.47\textwidth]{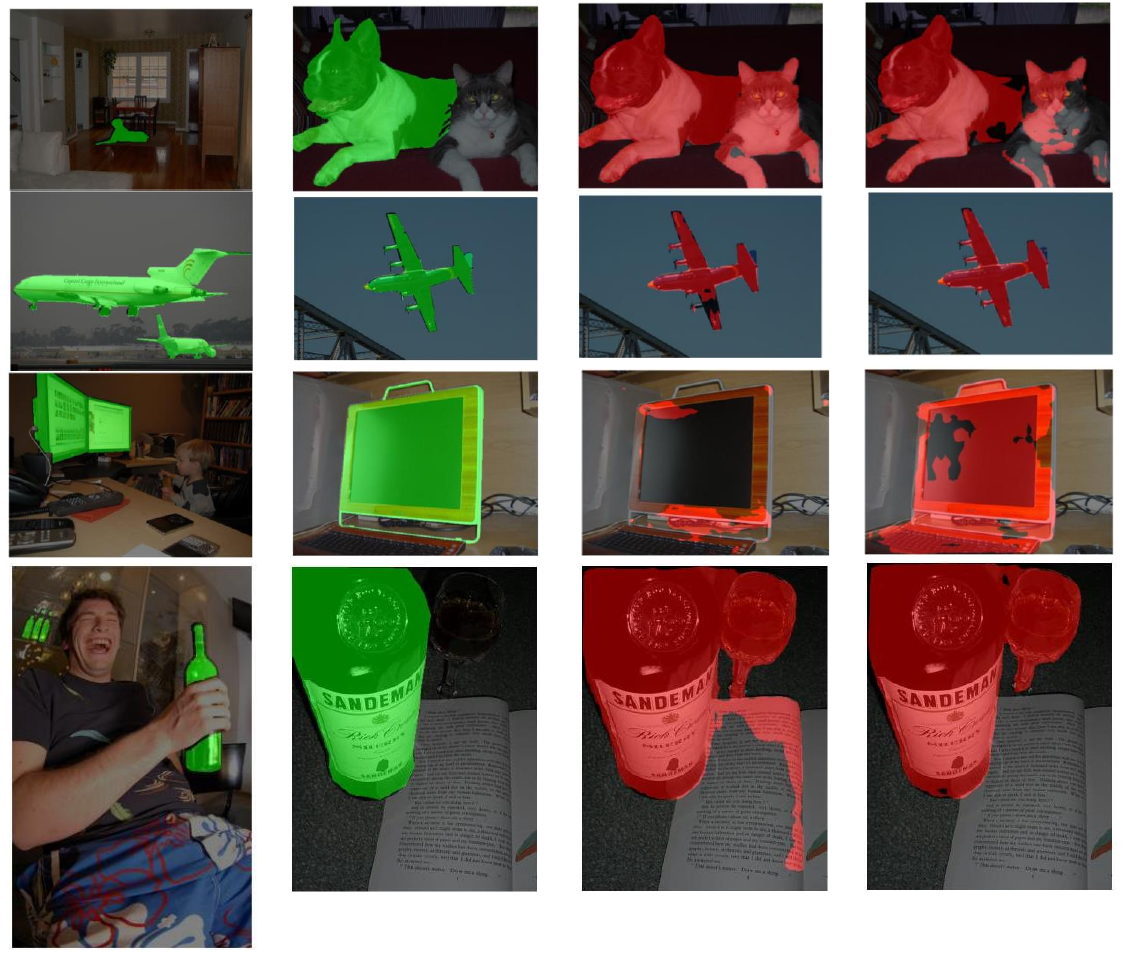}
\caption[Qualitative results]{Visualization of one-shot semantic segmentation.}
\label{fig:qualitative}
\end{figure}

In Fig. \ref{fig:qualitative}, we provided a detailed set of one-shot semantic segmentation results, including the support image, ground truth, and the predicted output before and after applying our Diffusion Synthesis.
Notably, our model demonstrates impressive performance using just a single support image for reference. Results before and after applying our Diffusion Synthesis method reveal a significant improvement in segmentation quality. In the first and third rows, the predicted masks align closely with the ground truth, showing highly accurate predictions. Even in more scenarios, such as men riding bicycles where objects are intertwined as seen in Fig. \ref{fig:qualitative}, our method significantly enhances segmentation performance. More results are presented in supplementary material.


\subsection{Comparison with State-of-the-Art Methods}
\label{ssec:comparison}
\begin{figure}[t]
\centering\includegraphics[width=0.45\textwidth]{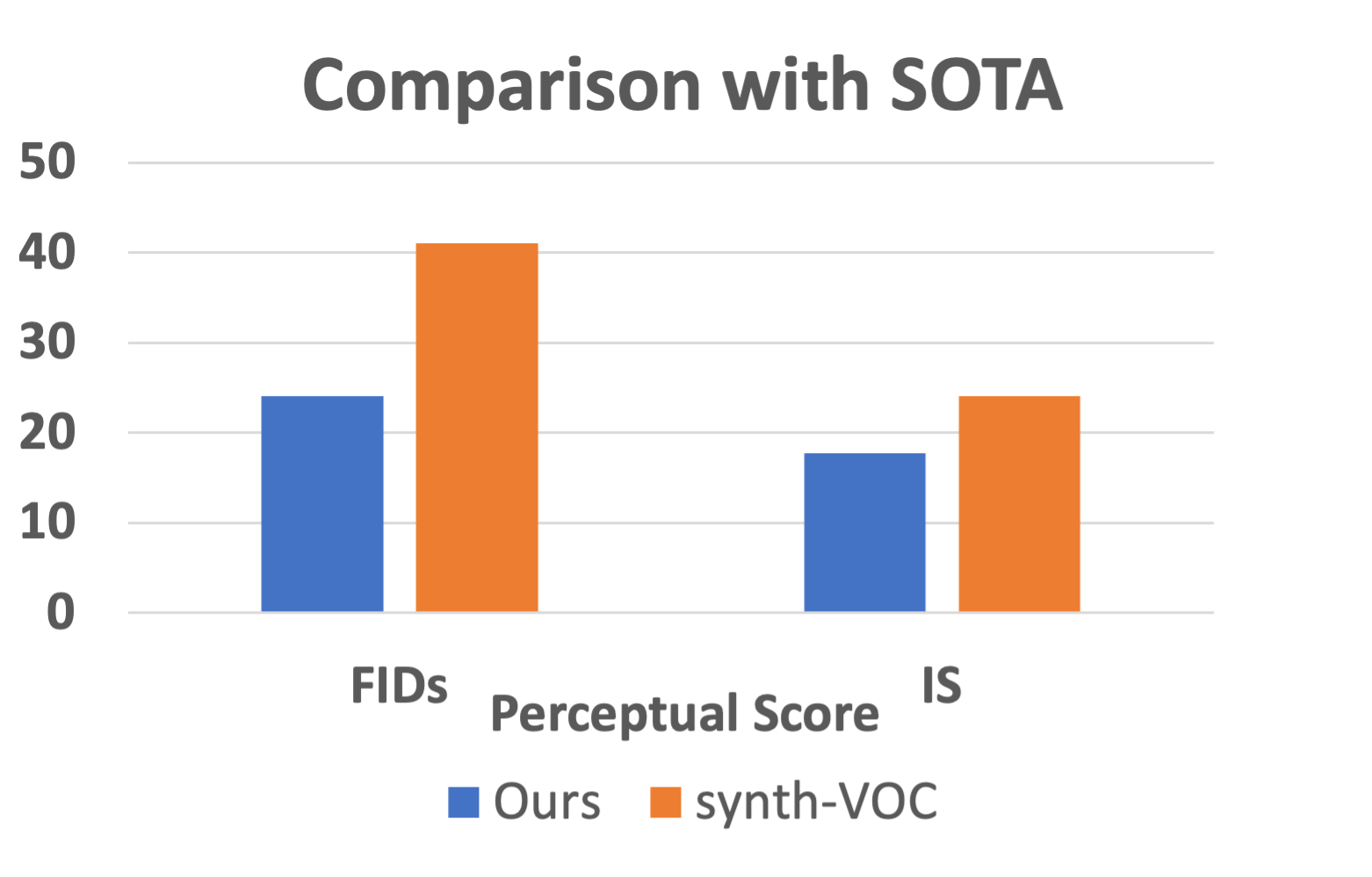}
\caption[Qualitative results]{Synthetic dataset performance comparison between ours and synth-VOC.}
\label{fig:sota}
\end{figure}

We compare our method with the state-of-the-art model-based dataset synthesis approach, Dataset Diffusion \cite{nguyen2023dataset}, which leverages a pre-trained diffusion model to generate synthetic data for semantic segmentation. To ensure a fair comparison, we evaluate one-shot semantic segmentation using the provided synthetic dataset, synth-VOC, and compare the results with our method. The benchmark dataset, synth-VOC, consists of 44,430 images and corresponding masks. We conduct augmentation experiments with a synthetic probability of $\alpha=0.4$ for the one-shot semantic segmentation models BAM and HDMNet, following the same training setup described in the previous section. As shown in Table \ref{table:fsspascal} and Fig. \ref{fig:sota}, our synthetic dataset outperforms Dataset Diffusion, achieving a 15.92 lower FID, and a 1.4\% and 1.7\% higher mIoU for HDMNet and BAM in the one-shot semantic segmentation, respectively. This improvement can be attributed to our high-fidelity generation, which requires minimal human effort and is guided by carefully designed prompts, selection and alignment processes. While our data shows slightly lower IS results, this could be due to challenges in learning underrepresented classes, leading to a bias towards dominant classes. Overall, the results demonstrate that our synthetic dataset not only enhances perceptual quality but also significantly improves performance in downstream task.

\subsection{Ablation Study}
\label{ssec:ablation}

\subsubsection{Ablation of Multi-way Prompt Generator.}
\label{ssec:prompt}
Table\ref{tab:prompt} presents the evaluation of different prompt designs in our MPG, including (1) class prompts (2) captions (3) our prompt approach. Specifically, our prompt design technique increases the perceptual performance by 1.93 and 8.66 FID, 1.28 and 5.13 IS over the ``caption-only" and ``class-label-only" text prompts, respectively. 
A comparison of synthetic data generated by different prompt designs is illustrated in Fig. \ref{fig:promptablation}. These visual examples clearly demonstrate the significant improvement in image quality achieved by the MPG. Specifically, our approach leads to more detailed and contextually accurate images, which are essential for enhancing downstream task performance. 

\begin{table}[]
    \centering
    \begin{tabular}{cccc}
        \hline
       \textbf{Type} & \textbf{Class} & \textbf{Caption} & \textbf{Our prompt } \\
       \hline
       FID$\downarrow$  & 33.43 & 26.02 & \textbf{24.09} \\
       IS$\uparrow$ & 12.60 & 16.45 & \textbf{17.73} \\
       \hline
    \end{tabular}
    \caption{Ablation study of Prompt. 
    Quality of synthetic data on PASCAL-${5^i}$.}
    \label{tab:prompt}
\end{table}

\subsubsection{Ablation of synthesis probability.}
\label{ssec:synthesisprobability} 
\begin{figure}[t]
\centering\includegraphics[width=0.23\textwidth]{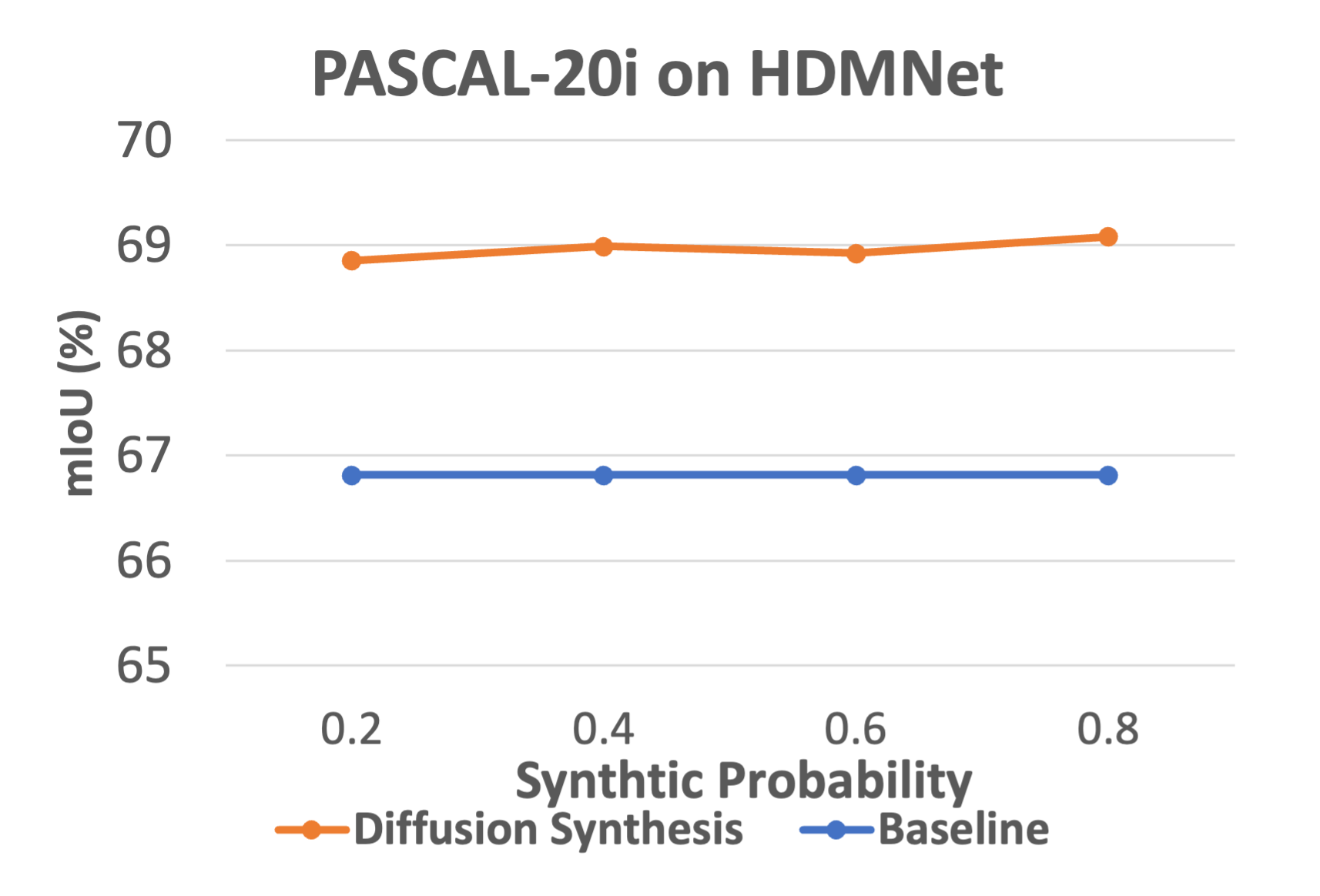} 
\centering\includegraphics[width=0.23\textwidth]{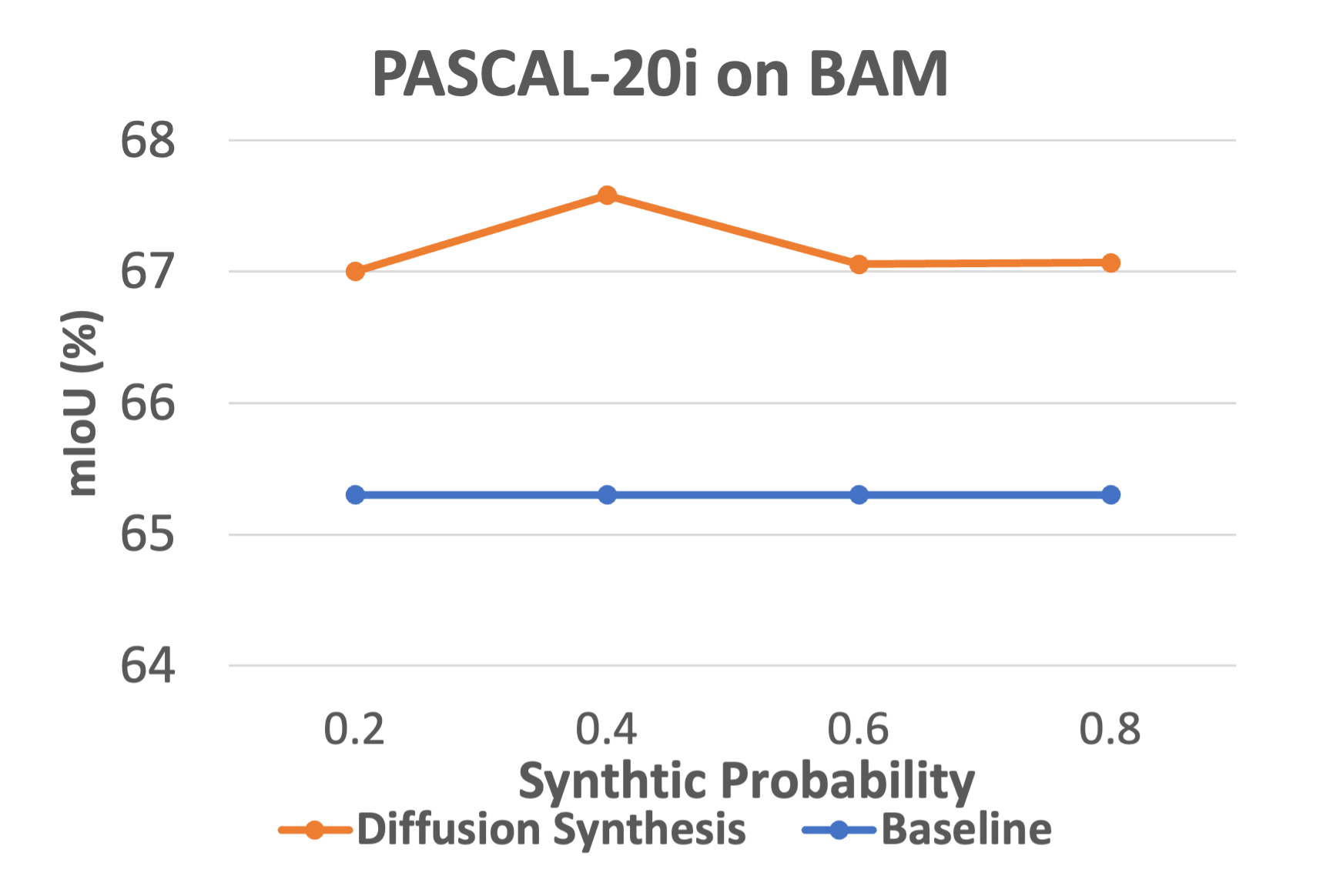} 
\caption[Diffusion-Synthesis module framework.]{Ablation of synthesis probability on PASCAL-$5^i$.}
\label{fig:ratio_ablation}
\end{figure}

We next conduct an ablation study to analyze the sensitivity of our method to the ratio of real and synthetic data, controlled by the hyperparameter $\alpha \in [0, 1]$, which represents the probability of sampling synthetic images during training. As shown in Fig. \ref{fig:ratio_ablation},  illustrates that our method consistently enhances performance, culminating in a substantial improvement of 2.2\% at $\alpha = 0.4$ for BAM. This indicates that our approach enhances the performance on downstream tasks steadily, as increasing sampling frequency does not deteriorate performance a lot. 


\section{Conclusion and Discussion}
\label{sec:conclusion}

In this study, we introduced several key innovations in diffusion-based data generation. First, we proposed an automatic, architecture-agnostic data generation pipeline, denoted as Diffusion Synthesis, that leverages pre-trained Vision-Language Models and diffusion models. This pipeline operates without requiring ground-truth labels and requires minimal human intervention, significantly streamlining the data generation process. Second, we developed a  Multi-way Prompt Generator, Mask Generator and High-quality Image Selection module, to ensure the realism and high fidelity of synthetic data. This module effectively guarantees that generated images closely resemble real-world data. Finally, our approach resulted in significant performance improvements in downstream few-shot semantic segmentation tasks, narrowing the performance gap between few-shot and fully-supervised models. 
These advancements collectively represent a major step forward in efficiently generating high-quality synthetic data and enhancing machine learning model performance in low-data scenarios.

\subsection{Limitations}
However, two challenges need be addressed to fully harness potential of diffusion model-based data synthesis in the future. First, the inherent biases in multi-modal large models, pre-trained on vast internet-scale datasets, can propagate into the synthetic data, potentially skewing the outcomes of downstream applications. Second, while diffusion models have shown effectiveness in data augmentation, their utility is highly dependent on the specific dataset and task, with few-shot learning being particularly sensitive. Future research should focus on optimizing these models for task-specific applications, tailoring augmentation strategies to enhance performance in low-data scenarios.

\bibliographystyle{aaai25}
\bibliography{aaai25}

\begin{thebibliography}{67}
\providecommand{\natexlab}[1]{#1}

\bibitem[{Ali et~al.(2019)Ali, Salleh, Saedudin, Hussain, and Mushtaq}]{article}
Ali, H.; Salleh, M.; Saedudin, R.; Hussain, K.; and Mushtaq, M. 2019.
\newblock Imbalance class problems in data mining: A review.
\newblock \emph{Indonesian Journal of Electrical Engineering and Computer Science}, 14.

\bibitem[{Ali-Gombe and Elyan(2019)}]{ali2019mfc}
Ali-Gombe, A.; and Elyan, E. 2019.
\newblock MFC-GAN: Class-imbalanced dataset classification using multiple fake class generative adversarial network.
\newblock \emph{Neurocomputing}, 361: 212--221.

\bibitem[{Bochkovskiy, Wang, and Liao(2020)}]{bochkovskiy2020yolov4}
Bochkovskiy, A.; Wang, C.-Y.; and Liao, H.-Y.~M. 2020.
\newblock Yolov4: Optimal speed and accuracy of object detection.
\newblock \emph{arXiv preprint arXiv:2004.10934}.

\bibitem[{Chen et~al.(2017)Chen, Papandreou, Schroff, and Adam}]{chen2017rethinking}
Chen, L.-C.; Papandreou, G.; Schroff, F.; and Adam, H. 2017.
\newblock Rethinking atrous convolution for semantic image segmentation.
\newblock \emph{arXiv preprint arXiv:1706.05587}.

\bibitem[{Cheng et~al.(2022)Cheng, Misra, Schwing, Kirillov, and Girdhar}]{cheng2022masked}
Cheng, B.; Misra, I.; Schwing, A.~G.; Kirillov, A.; and Girdhar, R. 2022.
\newblock Masked-attention mask transformer for universal image segmentation.
\newblock In \emph{Proceedings of the IEEE/CVF conference on computer vision and pattern recognition}, 1290--1299.

\bibitem[{Choi et~al.(2023)Choi, Choi, Kim, Kim, and Yoon}]{choi2023custom}
Choi, J.; Choi, Y.; Kim, Y.; Kim, J.; and Yoon, S. 2023.
\newblock Custom-edit: Text-guided image editing with customized diffusion models.
\newblock \emph{arXiv preprint arXiv:2305.15779}.

\bibitem[{Cordts et~al.(2016)Cordts, Omran, Ramos, Rehfeld, Enzweiler, Benenson, Franke, Roth, and Schiele}]{cordts2016cityscapes}
Cordts, M.; Omran, M.; Ramos, S.; Rehfeld, T.; Enzweiler, M.; Benenson, R.; Franke, U.; Roth, S.; and Schiele, B. 2016.
\newblock The cityscapes dataset for semantic urban scene understanding.
\newblock In \emph{Proceedings of the IEEE conference on computer vision and pattern recognition}, 3213--3223.

\bibitem[{Couairon et~al.(2022)Couairon, Verbeek, Schwenk, and Cord}]{couairon2022diffedit}
Couairon, G.; Verbeek, J.; Schwenk, H.; and Cord, M. 2022.
\newblock Diffedit: Diffusion-based semantic image editing with mask guidance.
\newblock \emph{arXiv preprint arXiv:2210.11427}.

\bibitem[{Dabouei et~al.(2021)Dabouei, Soleymani, Taherkhani, and Nasrabadi}]{dabouei2021supermix}
Dabouei, A.; Soleymani, S.; Taherkhani, F.; and Nasrabadi, N.~M. 2021.
\newblock Supermix: Supervising the mixing data augmentation.
\newblock In \emph{Proceedings of the IEEE/CVF conference on computer vision and pattern recognition}, 13794--13803.

\bibitem[{Dong et~al.(2023)Dong, Xue, Duan, and Han}]{dong2023prompt}
Dong, W.; Xue, S.; Duan, X.; and Han, S. 2023.
\newblock Prompt tuning inversion for text-driven image editing using diffusion models.
\newblock In \emph{Proceedings of the IEEE/CVF International Conference on Computer Vision}, 7430--7440.

\bibitem[{Dosovitskiy et~al.(2020)Dosovitskiy, Beyer, Kolesnikov, Weissenborn, Zhai, Unterthiner, Dehghani, Minderer, Heigold, Gelly et~al.}]{dosovitskiy2020image}
Dosovitskiy, A.; Beyer, L.; Kolesnikov, A.; Weissenborn, D.; Zhai, X.; Unterthiner, T.; Dehghani, M.; Minderer, M.; Heigold, G.; Gelly, S.; et~al. 2020.
\newblock An image is worth 16x16 words: Transformers for image recognition at scale.
\newblock \emph{arXiv preprint arXiv:2010.11929}.

\bibitem[{Douzas and Bacao(2018)}]{douzas2018effective}
Douzas, G.; and Bacao, F. 2018.
\newblock Effective data generation for imbalanced learning using conditional generative adversarial networks.
\newblock \emph{Expert Systems with applications}, 91: 464--471.

\bibitem[{Dvornik, Mairal, and Schmid(2018)}]{dvornik2018modeling}
Dvornik, N.; Mairal, J.; and Schmid, C. 2018.
\newblock Modeling visual context is key to augmenting object detection datasets.
\newblock In \emph{Proceedings of the European Conference on Computer Vision (ECCV)}, 364--380.

\bibitem[{Dwibedi, Misra, and Hebert(2017)}]{dwibedi2017cut}
Dwibedi, D.; Misra, I.; and Hebert, M. 2017.
\newblock Cut, paste and learn: Surprisingly easy synthesis for instance detection.
\newblock In \emph{Proceedings of the IEEE international conference on computer vision}, 1301--1310.

\bibitem[{Everingham et~al.(2012)Everingham, Van~Gool, Williams, Winn, and Zisserman}]{pascal-voc-2012}
Everingham, M.; Van~Gool, L.; Williams, C. K.~I.; Winn, J.; and Zisserman, A. 2012.
\newblock The {PASCAL} {V}isual {O}bject {C}lasses {C}hallenge 2012 {(VOC2012)} {R}esults.
\newblock http://www.pascal-network.org/challenges/VOC/voc2012/workshop/index.html.

\bibitem[{Fan et~al.(2022)Fan, Pei, Tai, and Tang}]{fan2022ssp}
Fan, Q.; Pei, W.; Tai, Y.-W.; and Tang, C.-K. 2022.
\newblock Self-Support Few-Shot Semantic Segmentation.

\bibitem[{Feng et~al.(2023)Feng, Yu, Liu, Khan, and Zuo}]{feng2023diverse}
Feng, C.-M.; Yu, K.; Liu, Y.; Khan, S.; and Zuo, W. 2023.
\newblock Diverse data augmentation with diffusions for effective test-time prompt tuning.
\newblock In \emph{Proceedings of the IEEE/CVF International Conference on Computer Vision}, 2704--2714.

\bibitem[{Georgakis et~al.(2017)Georgakis, Mousavian, Berg, and Kosecka}]{georgakis2017synthesizing}
Georgakis, G.; Mousavian, A.; Berg, A.~C.; and Kosecka, J. 2017.
\newblock Synthesizing training data for object detection in indoor scenes.
\newblock \emph{arXiv preprint arXiv:1702.07836}.

\bibitem[{Ghiasi et~al.(2021)Ghiasi, Cui, Srinivas, Qian, Lin, Cubuk, Le, and Zoph}]{ghiasi2021simple}
Ghiasi, G.; Cui, Y.; Srinivas, A.; Qian, R.; Lin, T.-Y.; Cubuk, E.~D.; Le, Q.~V.; and Zoph, B. 2021.
\newblock Simple copy-paste is a strong data augmentation method for instance segmentation.
\newblock In \emph{Proceedings of the IEEE/CVF conference on computer vision and pattern recognition}, 2918--2928.

\bibitem[{Hariharan et~al.(2011)Hariharan, Arbeláez, Bourdev, Maji, and Malik}]{6126343}
Hariharan, B.; Arbeláez, P.; Bourdev, L.; Maji, S.; and Malik, J. 2011.
\newblock Semantic contours from inverse detectors.
\newblock In \emph{2011 International Conference on Computer Vision}, 991--998.

\bibitem[{Hashimoto et~al.(2020)Hashimoto, Fukushima, Koga, Takagi, Ko, Kohno, Nakaguro, Nakamura, Hontani, and Takeuchi}]{hashimoto2020multi}
Hashimoto, N.; Fukushima, D.; Koga, R.; Takagi, Y.; Ko, K.; Kohno, K.; Nakaguro, M.; Nakamura, S.; Hontani, H.; and Takeuchi, I. 2020.
\newblock Multi-scale domain-adversarial multiple-instance CNN for cancer subtype classification with unannotated histopathological images.
\newblock In \emph{Proceedings of the IEEE/CVF conference on computer vision and pattern recognition}, 3852--3861.

\bibitem[{He et~al.(2016)He, Zhang, Ren, and Sun}]{he2016deep}
He, K.; Zhang, X.; Ren, S.; and Sun, J. 2016.
\newblock Deep residual learning for image recognition.
\newblock In \emph{Proceedings of the IEEE conference on computer vision and pattern recognition}, 770--778.

\bibitem[{He et~al.(2022)He, Sun, Yu, Xue, Zhang, Torr, Bai, and Qi}]{he2022synthetic}
He, R.; Sun, S.; Yu, X.; Xue, C.; Zhang, W.; Torr, P.; Bai, S.; and Qi, X. 2022.
\newblock Is synthetic data from generative models ready for image recognition?
\newblock \emph{arXiv preprint arXiv:2210.07574}.

\bibitem[{Heusel et~al.(2017)Heusel, Ramsauer, Unterthiner, Nessler, and Hochreiter}]{heusel2017gans}
Heusel, M.; Ramsauer, H.; Unterthiner, T.; Nessler, B.; and Hochreiter, S. 2017.
\newblock Gans trained by a two time-scale update rule converge to a local nash equilibrium.
\newblock \emph{Advances in neural information processing systems}, 30.

\bibitem[{Ho, Jain, and Abbeel(2020)}]{ho2020denoising}
Ho, J.; Jain, A.; and Abbeel, P. 2020.
\newblock Denoising diffusion probabilistic models.
\newblock \emph{Advances in neural information processing systems}, 33: 6840--6851.

\bibitem[{Hong, Choi, and Kim(2021)}]{hong2021stylemix}
Hong, M.; Choi, J.; and Kim, G. 2021.
\newblock Stylemix: Separating content and style for enhanced data augmentation.
\newblock In \emph{Proceedings of the IEEE/CVF conference on computer vision and pattern recognition}, 14862--14870.

\bibitem[{Huang et~al.(2024{\natexlab{a}})Huang, Huang, Liu, Yan, Lv, Liu, Xiong, Zhang, Chen, and Cao}]{huang2024diffusion}
Huang, Y.; Huang, J.; Liu, Y.; Yan, M.; Lv, J.; Liu, J.; Xiong, W.; Zhang, H.; Chen, S.; and Cao, L. 2024{\natexlab{a}}.
\newblock Diffusion model-based image editing: A survey.
\newblock \emph{arXiv preprint arXiv:2402.17525}.

\bibitem[{Huang et~al.(2024{\natexlab{b}})Huang, Xie, Wang, Yuan, Cun, Ge, Zhou, Dong, Huang, Zhang et~al.}]{huang2024smartedit}
Huang, Y.; Xie, L.; Wang, X.; Yuan, Z.; Cun, X.; Ge, Y.; Zhou, J.; Dong, C.; Huang, R.; Zhang, R.; et~al. 2024{\natexlab{b}}.
\newblock Smartedit: Exploring complex instruction-based image editing with multimodal large language models.
\newblock In \emph{Proceedings of the IEEE/CVF Conference on Computer Vision and Pattern Recognition}, 8362--8371.

\bibitem[{Inoue(2018)}]{inoue2018data}
Inoue, H. 2018.
\newblock Data augmentation by pairing samples for images classification.
\newblock \emph{arXiv preprint arXiv:1801.02929}.

\bibitem[{Kirillov et~al.(2023)Kirillov, Mintun, Ravi, Mao, Rolland, Gustafson, Xiao, Whitehead, Berg, Lo et~al.}]{kirillov2023segment}
Kirillov, A.; Mintun, E.; Ravi, N.; Mao, H.; Rolland, C.; Gustafson, L.; Xiao, T.; Whitehead, S.; Berg, A.~C.; Lo, W.-Y.; et~al. 2023.
\newblock Segment anything.
\newblock In \emph{Proceedings of the IEEE/CVF International Conference on Computer Vision}, 4015--4026.

\bibitem[{Lang et~al.(2022)Lang, Cheng, Tu, and Han}]{lang2022bam}
Lang, C.; Cheng, G.; Tu, B.; and Han, J. 2022.
\newblock Learning What Not to Segment: A New Perspective on Few-Shot Segmentation.
\newblock In \emph{Proceedings of the IEEE/CVF Conference on Computer Vision and Pattern Recognition (CVPR)}, 8057--8067.

\bibitem[{Li et~al.(2022{\natexlab{a}})Li, Ling, Kim, Kreis, Fidler, and Torralba}]{li2022bigdatasetgan}
Li, D.; Ling, H.; Kim, S.~W.; Kreis, K.; Fidler, S.; and Torralba, A. 2022{\natexlab{a}}.
\newblock Bigdatasetgan: Synthesizing imagenet with pixel-wise annotations.
\newblock In \emph{Proceedings of the IEEE/CVF Conference on Computer Vision and Pattern Recognition}, 21330--21340.

\bibitem[{Li et~al.(2022{\natexlab{b}})Li, Li, Xiong, and Hoi}]{li2022blip}
Li, J.; Li, D.; Xiong, C.; and Hoi, S. 2022{\natexlab{b}}.
\newblock Blip: Bootstrapping language-image pre-training for unified vision-language understanding and generation.
\newblock In \emph{International conference on machine learning}, 12888--12900. PMLR.

\bibitem[{Li et~al.(2020)Li, Yu, Tan, Mei, Tang, Shen, Yuille, and Xie}]{li2020shape}
Li, Y.; Yu, Q.; Tan, M.; Mei, J.; Tang, P.; Shen, W.; Yuille, A.; and Xie, C. 2020.
\newblock Shape-texture debiased neural network training.
\newblock \emph{arXiv preprint arXiv:2010.05981}, 2.

\bibitem[{Lin et~al.(2014)Lin, Maire, Belongie, Hays, Perona, Ramanan, Doll{\'a}r, and Zitnick}]{lin2014microsoft}
Lin, T.-Y.; Maire, M.; Belongie, S.; Hays, J.; Perona, P.; Ramanan, D.; Doll{\'a}r, P.; and Zitnick, C.~L. 2014.
\newblock Microsoft coco: Common objects in context.
\newblock In \emph{Computer Vision--ECCV 2014: 13th European Conference, Zurich, Switzerland, September 6-12, 2014, Proceedings, Part V 13}, 740--755. Springer.

\bibitem[{Liu et~al.(2023)Liu, Zhu, Li, Chen, Wang, and Shen}]{liu2023matcher}
Liu, Y.; Zhu, M.; Li, H.; Chen, H.; Wang, X.; and Shen, C. 2023.
\newblock Matcher: Segment anything with one shot using all-purpose feature matching.
\newblock \emph{arXiv preprint arXiv:2305.13310}.

\bibitem[{Mandi et~al.(2022)Mandi, Bharadhwaj, Moens, Song, Rajeswaran, and Kumar}]{mandi2022cacti}
Mandi, Z.; Bharadhwaj, H.; Moens, V.; Song, S.; Rajeswaran, A.; and Kumar, V. 2022.
\newblock Cacti: A framework for scalable multi-task multi-scene visual imitation learning.
\newblock \emph{arXiv preprint arXiv:2212.05711}.

\bibitem[{Mokady et~al.(2023)Mokady, Hertz, Aberman, Pritch, and Cohen-Or}]{mokady2023null}
Mokady, R.; Hertz, A.; Aberman, K.; Pritch, Y.; and Cohen-Or, D. 2023.
\newblock Null-text inversion for editing real images using guided diffusion models.
\newblock In \emph{Proceedings of the IEEE/CVF Conference on Computer Vision and Pattern Recognition}, 6038--6047.

\bibitem[{Nguyen and Todorovic(2019)}]{nguyen2019feature}
Nguyen, K.; and Todorovic, S. 2019.
\newblock Feature weighting and boosting for few-shot segmentation.
\newblock In \emph{Proceedings of the IEEE/CVF International Conference on Computer Vision}, 622--631.

\bibitem[{Nguyen et~al.(2023)Nguyen, Vu, Tran, and Nguyen}]{nguyen2023dataset}
Nguyen, Q.~H.; Vu, T.~T.; Tran, A.~T.; and Nguyen, K. 2023.
\newblock Dataset Diffusion: Diffusion-based Synthetic Data Generation for Pixel-Level Semantic Segmentation.
\newblock In \emph{Thirty-seventh Conference on Neural Information Processing Systems}.

\bibitem[{Nichol and Dhariwal(2021)}]{nichol2021improved}
Nichol, A.~Q.; and Dhariwal, P. 2021.
\newblock Improved denoising diffusion probabilistic models.
\newblock In \emph{International conference on machine learning}, 8162--8171. PMLR.

\bibitem[{Pan et~al.(2023)Pan, Gherardi, Xie, and Huang}]{pan2023effective}
Pan, Z.; Gherardi, R.; Xie, X.; and Huang, S. 2023.
\newblock Effective real image editing with accelerated iterative diffusion inversion.
\newblock In \emph{Proceedings of the IEEE/CVF International Conference on Computer Vision}, 15912--15921.

\bibitem[{Peng et~al.(2023)Peng, Tian, Wu, Wang, Liu, Su, and Jia}]{peng2023hierarchical}
Peng, B.; Tian, Z.; Wu, X.; Wang, C.; Liu, S.; Su, J.; and Jia, J. 2023.
\newblock Hierarchical Dense Correlation Distillation for Few-Shot Segmentation.
\newblock \emph{arXiv preprint arXiv:2303.14652}.

\bibitem[{Rombach et~al.(2021)Rombach, Blattmann, Lorenz, Esser, and Ommer}]{rombach2021highresolution}
Rombach, R.; Blattmann, A.; Lorenz, D.; Esser, P.; and Ommer, B. 2021.
\newblock High-Resolution Image Synthesis with Latent Diffusion Models.
\newblock arXiv:2112.10752.

\bibitem[{Salimans et~al.(2016)Salimans, Goodfellow, Zaremba, Cheung, Radford, and Chen}]{salimans2016improved}
Salimans, T.; Goodfellow, I.; Zaremba, W.; Cheung, V.; Radford, A.; and Chen, X. 2016.
\newblock Improved techniques for training gans.
\newblock \emph{Advances in neural information processing systems}, 29.

\bibitem[{Shaban et~al.(2017)Shaban, Bansal, Liu, Essa, and Boots}]{shaban2017one}
Shaban, A.; Bansal, S.; Liu, Z.; Essa, I.; and Boots, B. 2017.
\newblock One-shot learning for semantic segmentation.
\newblock \emph{arXiv preprint arXiv:1709.03410}.

\bibitem[{Shi et~al.(2022)Shi, Wei, Zhang, Lu, Ning, Chen, Ma, and Zheng}]{shi2022dense}
Shi, X.; Wei, D.; Zhang, Y.; Lu, D.; Ning, M.; Chen, J.; Ma, K.; and Zheng, Y. 2022.
\newblock Dense Cross-Query-and-Support Attention Weighted Mask Aggregation for Few-Shot Segmentation.
\newblock In \emph{European Conference on Computer Vision}, 151--168. Springer.

\bibitem[{Song, Meng, and Ermon(2020)}]{song2020denoising}
Song, J.; Meng, C.; and Ermon, S. 2020.
\newblock Denoising diffusion implicit models.
\newblock \emph{arXiv preprint arXiv:2010.02502}.

\bibitem[{Tan, Chen, and Yan(2023)}]{tan2023diffss}
Tan, W.; Chen, S.; and Yan, B. 2023.
\newblock Diffss: Diffusion model for few-shot semantic segmentation.
\newblock \emph{arXiv preprint arXiv:2307.00773}.

\bibitem[{Trabucco et~al.(2023)Trabucco, Doherty, Gurinas, and Salakhutdinov}]{trabucco2023effective}
Trabucco, B.; Doherty, K.; Gurinas, M.; and Salakhutdinov, R. 2023.
\newblock Effective data augmentation with diffusion models.
\newblock \emph{arXiv preprint arXiv:2302.07944}.

\bibitem[{Tran et~al.(2017)Tran, Pham, Carneiro, Palmer, and Reid}]{tran2017bayesian}
Tran, T.; Pham, T.; Carneiro, G.; Palmer, L.; and Reid, I. 2017.
\newblock A bayesian data augmentation approach for learning deep models.
\newblock \emph{Advances in neural information processing systems}, 30.

\bibitem[{Wang, Zhao, and Xing(2023)}]{wang2023stylediffusion}
Wang, Z.; Zhao, L.; and Xing, W. 2023.
\newblock Stylediffusion: Controllable disentangled style transfer via diffusion models.
\newblock In \emph{Proceedings of the IEEE/CVF International Conference on Computer Vision}, 7677--7689.

\bibitem[{Willemink et~al.(2020)Willemink, Koszek, Hardell, Wu, Fleischmann, Harvey, Folio, Summers, Rubin, and Lungren}]{Willemink2020PreparingMI}
Willemink, M.~J.; Koszek, W.~A.; Hardell, C.; Wu, J.; Fleischmann, D.; Harvey, H.; Folio, L.~R.; Summers, R.~M.; Rubin, D.; and Lungren, M.~P. 2020.
\newblock Preparing Medical Imaging Data for Machine Learning.
\newblock \emph{Radiology}, 192224.

\bibitem[{Wu et~al.(2023)Wu, Zhao, Shou, Zhou, and Shen}]{wu2023diffumask}
Wu, W.; Zhao, Y.; Shou, M.~Z.; Zhou, H.; and Shen, C. 2023.
\newblock Diffumask: Synthesizing images with pixel-level annotations for semantic segmentation using diffusion models.
\newblock In \emph{Proceedings of the IEEE/CVF International Conference on Computer Vision}, 1206--1217.

\bibitem[{Xu et~al.(2022)Xu, Yoon, Fuentes, Yang, and Park}]{xu2022style}
Xu, M.; Yoon, S.; Fuentes, A.; Yang, J.; and Park, D.~S. 2022.
\newblock Style-consistent image translation: A novel data augmentation paradigm to improve plant disease recognition.
\newblock \emph{Frontiers in plant science}, 12: 773142.

\bibitem[{Xu et~al.(2023)Xu, Zhao, Lin, and Long}]{xu2023self}
Xu, Q.; Zhao, W.; Lin, G.; and Long, C. 2023.
\newblock Self-calibrated cross attention network for few-shot segmentation.
\newblock In \emph{Proceedings of the IEEE/CVF International Conference on Computer Vision}, 655--665.

\bibitem[{Xu et~al.(2024)Xu, Ma, Huang, Lee, and Chai}]{xu2024cyclenet}
Xu, S.; Ma, Z.; Huang, Y.; Lee, H.; and Chai, J. 2024.
\newblock Cyclenet: Rethinking cycle consistency in text-guided diffusion for image manipulation.
\newblock \emph{Advances in Neural Information Processing Systems}, 36.

\bibitem[{Yang et~al.(2023)Yang, Chen, Feng, and Huang}]{yang2023mianet}
Yang, Y.; Chen, Q.; Feng, Y.; and Huang, T. 2023.
\newblock MIANet: Aggregating unbiased instance and general information for few-shot semantic segmentation.
\newblock In \emph{Proceedings of the IEEE/CVF Conference on Computer Vision and Pattern Recognition}, 7131--7140.

\bibitem[{Yun et~al.(2019)Yun, Han, Oh, Chun, Choe, and Yoo}]{yun2019cutmix}
Yun, S.; Han, D.; Oh, S.~J.; Chun, S.; Choe, J.; and Yoo, Y. 2019.
\newblock Cutmix: Regularization strategy to train strong classifiers with localizable features.
\newblock In \emph{Proceedings of the IEEE/CVF international conference on computer vision}, 6023--6032.

\bibitem[{Zhang et~al.(2021{\natexlab{a}})Zhang, Kang, Yang, and Wei}]{zhang2021few}
Zhang, G.; Kang, G.; Yang, Y.; and Wei, Y. 2021{\natexlab{a}}.
\newblock Few-shot segmentation via cycle-consistent transformer.
\newblock \emph{Advances in Neural Information Processing Systems}, 34: 21984--21996.

\bibitem[{Zhang et~al.(2017)Zhang, Cisse, Dauphin, and Lopez-Paz}]{zhang2017mixup}
Zhang, H.; Cisse, M.; Dauphin, Y.~N.; and Lopez-Paz, D. 2017.
\newblock mixup: Beyond empirical risk minimization.
\newblock \emph{arXiv preprint arXiv:1710.09412}.

\bibitem[{Zhang et~al.(2023{\natexlab{a}})Zhang, Li, Liu, Zhang, Su, Zhu, Ni, and Shum}]{zhang2023dino}
Zhang, H.; Li, F.; Liu, S.; Zhang, L.; Su, H.; Zhu, J.; Ni, L.; and Shum, H.-Y. 2023{\natexlab{a}}.
\newblock {DINO}: {DETR} with Improved DeNoising Anchor Boxes for End-to-End Object Detection.
\newblock In \emph{The Eleventh International Conference on Learning Representations}.

\bibitem[{Zhang, Rao, and Agrawala(2023)}]{zhang2023adding}
Zhang, L.; Rao, A.; and Agrawala, M. 2023.
\newblock Adding Conditional Control to Text-to-Image Diffusion Models.

\bibitem[{Zhang et~al.(2021{\natexlab{b}})Zhang, Ling, Gao, Yin, Lafleche, Barriuso, Torralba, and Fidler}]{zhang2021datasetgan}
Zhang, Y.; Ling, H.; Gao, J.; Yin, K.; Lafleche, J.-F.; Barriuso, A.; Torralba, A.; and Fidler, S. 2021{\natexlab{b}}.
\newblock Datasetgan: Efficient labeled data factory with minimal human effort.
\newblock In \emph{Proceedings of the IEEE/CVF Conference on Computer Vision and Pattern Recognition}, 10145--10155.

\bibitem[{Zhang et~al.(2023{\natexlab{b}})Zhang, Han, Ghosh, Metaxas, and Ren}]{zhang2023sine}
Zhang, Z.; Han, L.; Ghosh, A.; Metaxas, D.~N.; and Ren, J. 2023{\natexlab{b}}.
\newblock Sine: Single image editing with text-to-image diffusion models.
\newblock In \emph{Proceedings of the IEEE/CVF Conference on Computer Vision and Pattern Recognition}, 6027--6037.

\bibitem[{Zhao et~al.(2017)Zhao, Shi, Qi, Wang, and Jia}]{zhao2017pyramid}
Zhao, H.; Shi, J.; Qi, X.; Wang, X.; and Jia, J. 2017.
\newblock Pyramid scene parsing network.
\newblock In \emph{Proceedings of the IEEE conference on computer vision and pattern recognition}, 2881--2890.

\bibitem[{Zheng et~al.(2021)Zheng, Yu, Wu, Zheng, Zheng, and Lee}]{zheng2021generative}
Zheng, Z.; Yu, Z.; Wu, Y.; Zheng, H.; Zheng, B.; and Lee, M. 2021.
\newblock Generative adversarial network with multi-branch discriminator for imbalanced cross-species image-to-image translation.
\newblock \emph{Neural Networks}, 141: 355--371.

\end{thebibliography}


\newpage
\clearpage
\section{Supplementary Material}

\subsection{Hyperparameters}

Our approach builds upon ControlNet's hyperparameters, while also introducing additional parameters to enhance the diversity of the synthetic images. The specific values of these hyperparameters are detailed in Table \ref{tab:hyperparameter}.

\begin{table}[t]
    \centering
    \setlength{\tabcolsep}{1mm}
    \begin{tabular}{l|r}
          \Xhline{2\arrayrulewidth}
          \textbf{Hyperparamter Name} & \textbf{Value} \\
          \hline 
          Synthetic Probability $\alpha$ & 0.5 \\
          Selection Threshold $\epsilon$ & 0.8 \\
          Synthetic Images Per Real $K$  & 5 \\
          ControlNet Checkpoint          & lllyasviel/sd-controlnet-seg \\
          Diffusion Guidance Scale       & 7.5 \\
          ControlNet Denoising Steps     & 50  \\
          SAM Checkpoint                 &   vit\_h  \\
          Dino Checkpoint                & groundingdino\_swint\_ogc.pth \\
          Segmentation Learning Rate     & 0.0001 \\
          Segmentation Batch Size        & 2,  4 \\
          \Xhline{2\arrayrulewidth}
    \end{tabular}
    \caption{Hyperparameters and their values.}
    \label{tab:hyperparameter}
\end{table}

\subsection{Experimental Environments}
Software and Hardware environment:
\begin{itemize}
    \item CUDA version: release 11.8
    \item PyTorch version: 1.13.1
    \item GPU: NVIDIA A10 24GB
    \item CPU: Intel(R) Xeon(R) Silver 4314 CPU $@$ 2.40GHz
\end{itemize}

\subsection{Comparison with Model-based Data Synthesis/Augmentation Methods}

\begin{figure}[t]
    \centering
    \includegraphics[width=0.95 \linewidth]{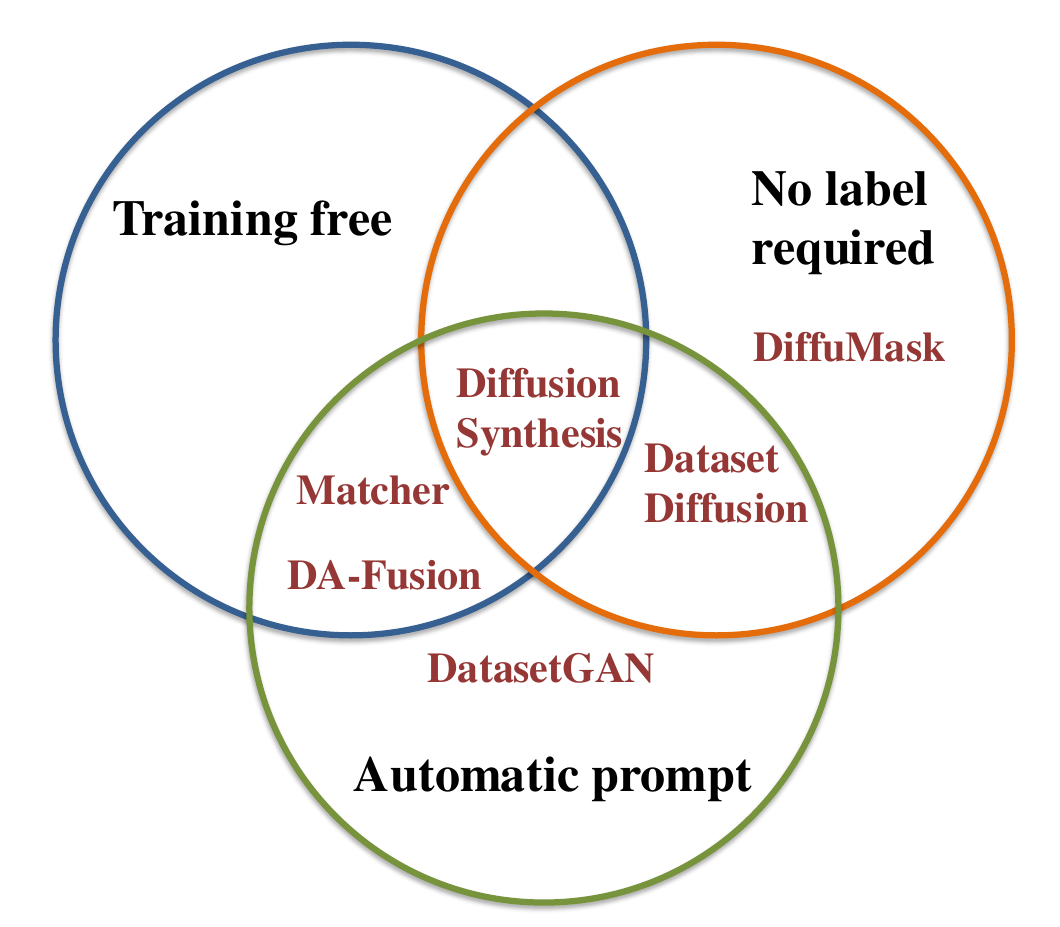}
    \caption{Comparison with state-of-the-art model-based data synthesis methods.}
    \label{fig:venn}
\end{figure}

We present a comprehensive comparison of model-based data synthesis and augmentation methods, as illustrated in Fig. \ref{fig:venn}. Our approach, Diffusion Synthesis, is evaluated alongside five other generative model-based techniques \cite{zhang2021datasetgan, nguyen2023dataset, wu2023diffumask, trabucco2023effective, liu2023matcher}, focusing on factors such as labeling effort, prompt engineering, and training requirements. The results of this rigorous comparison clearly demonstrate that Diffusion Synthesis generates high-quality image-mask pairs with minimal human intervention, significantly reducing the need for manual labeling and complex prompt engineering.

\subsection{Additional Results and Analysis}

\subsubsection{Ablation of High-quality Image Selection.}

\begin{figure}[t]
    \centering
    \includegraphics[width=0.95\linewidth]{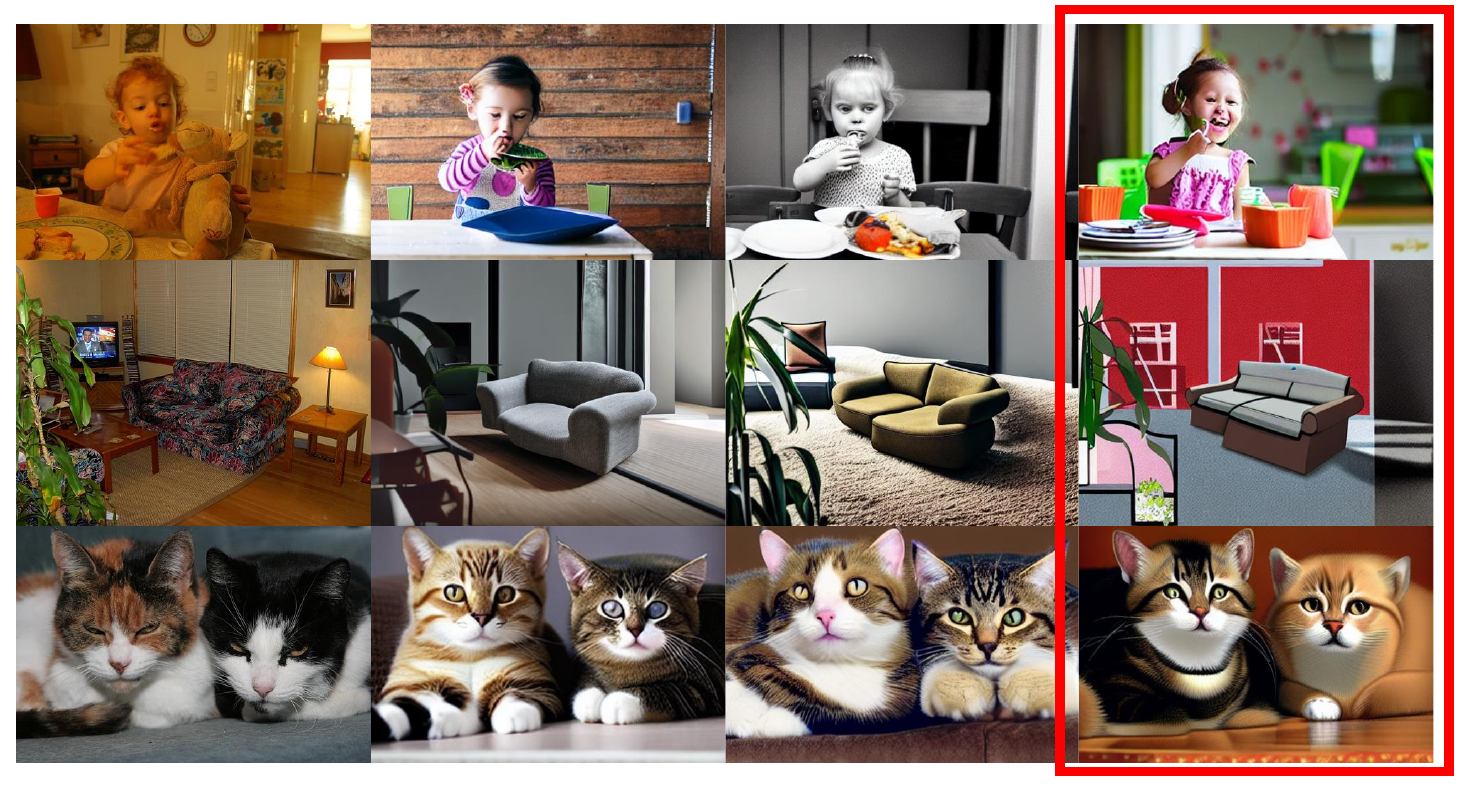}
    \caption{Visualization of image selection. From left to right: original image, three augmented images. The last column is the filtered out image.}
    \label{fig:his}
\end{figure}

We evaluate the effectiveness of our HIS module by visualizing augmented images and testing different filtration strengths. First, as illustrated in Fig. \ref{fig:his}, the HIS module successfully removes spurious augmentation. Thus, we can obtain augmented samples with diverse visual appearance variation while preserving realism. 

Additionally, we examine the impact of various similarity values of $\epsilon$ in Table \ref{tab:his}, to analyze the degree of diversity retained in the augmentations. The Table \ref{tab:his} demonstrates that our method achieves superior performance on $\epsilon=0.8$. This means that a certain amount of data is filtered out, indicating the high fidelity of synthetic data generated by our VLMs-based augmentation.

\begin{table}[]
    \centering
    \begin{tabular}{cccc}
        \Xhline{2\arrayrulewidth}
       \textbf{$\epsilon$} & FID$\downarrow$  & IS$\uparrow$ & mIoU$\uparrow$ \\
       \hline
       0.7 & 25.7 & 17.2 & 67.8 \\
       0.8 & 24.1 & 29.9 & 69.3 \\
       0.9 & 58.3 & 11.1 & 66.9 \\
       \hline
    \end{tabular}
    \caption{Ablation study of HIS of synthetic data on PASCAL-${5^i}$.}
    \label{tab:his}
\end{table}

\subsubsection{Results on fully-supervised semantic segmentation.}
Table \ref{tab:fs} presents a comparison of results of DeepLabV3 \cite{chen2017rethinking}, Mask2Former \cite{cheng2022masked} trained on various datasets: the real training set, the synthetic dataset generated by DiffuMask \cite{wu2023diffumask}, Dataset Diffusion \cite{nguyen2023dataset}, and our proposed Diffusion Synthesis. On the PASCAL VOC dataset, our approach achieves a competitive mIoU of 59.2, compared to the 79.9 mIoU of the real training set. Notably, our method significantly outperforms DiffuMusk by a large margin of 1.8 mIoU, using the same ResNet50 backbone. These results highlight the effectiveness of Diffusion Synthesis in generating high-quality synthetic data, though there remains a noticeable gap in performance when compared to models trained on real datasets. Furthermore, our approach demonstrates the ability to empower various semantic segmentation tasks.

\begin{table}[ht]
\centering
    \setlength{\tabcolsep}{1mm}
    \begin{tabular}{lll|rr}
    \Xhline{2\arrayrulewidth}
    \textbf{Segmenter} & \textbf{Dataset} & \textbf{Size} & \textbf{mIoU}  \\
    \hline
    DeepLabV3 & VOC's training & 11.5k & 77.4  \\
    Mask2Former & VOC's training & 11.5k & 77.3 \\
    Mask2Former & DiffuMask & 60k & 57.4 \\ 
    DeepLabV3 & Dataset Diffusion & 40.0k & 61.6 \\ 
    DeepLabV3 & Diffusion Synthesis & 16.5k & 59.2 \\
    \Xhline{2\arrayrulewidth}
    \end{tabular}
    \caption{Comparison in mIoU between training DeepLabV3 \cite{chen2017rethinking} and Mask2Former \cite{cheng2022masked} on the real training dataset, the synthetic dataset from DiffuMask \cite{wu2023diffumask}, Dataset Diffusion \cite{nguyen2023dataset}, and our Diffusion Synthesis.}
    \label{tab:fs}
\end{table}

    

\subsubsection{Additional results on COCO-$20^i$.}
More quantitative results on COCO-$20^i$ are presented in Table \ref{table:fsscoco_apendix}. Our augmentation method, Diffusion Synthesis, consistently boosts performance across all folds, with an average improvement of up to 1.2 $\%$ in mIoU for one-shot semantic segmentation on COCO-$20^i$. These results clearly demonstrate the effectiveness of our approach in enhancing model accuracy, particularly in challenging few-shot learning scenarios.

\subsubsection{More qualitative results.}
\begin{figure*}[t]
    \centering
    \includegraphics[width=0.9\linewidth]{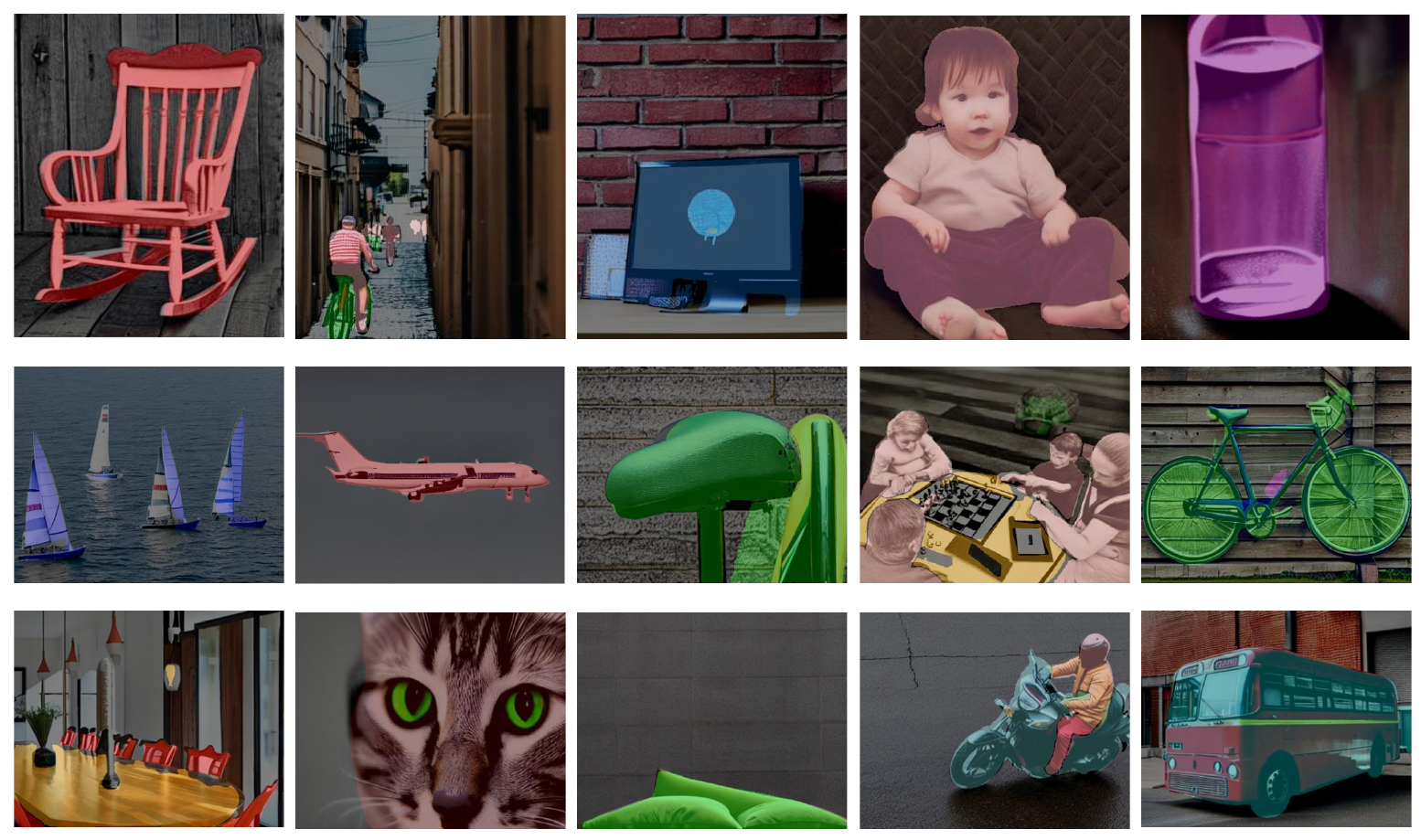}
    \caption{Synthesized image-mask pairs.}
    \label{fig:img-mask}
\end{figure*}

\begin{figure*}[t]
    \centering
    \hfill
    \begin{minipage}[t]{0.9\textwidth}
    \includegraphics[width=0.9\linewidth]{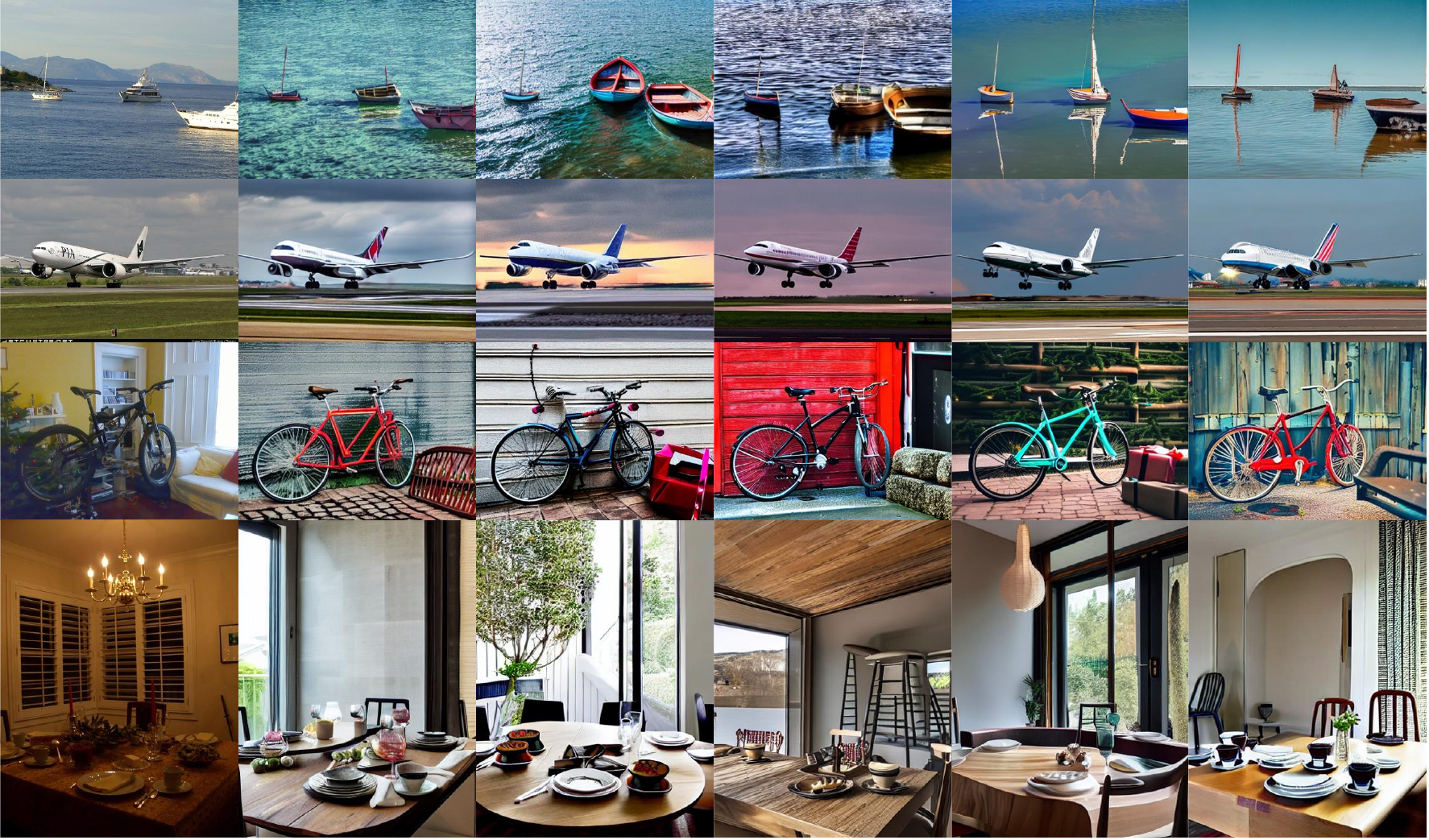}
    \caption*{(a). Synthesized data on PASCAL.}
    \centering
    \end{minipage}
    \begin{minipage}[t]{0.9\textwidth}
    \includegraphics[width=0.9\linewidth]{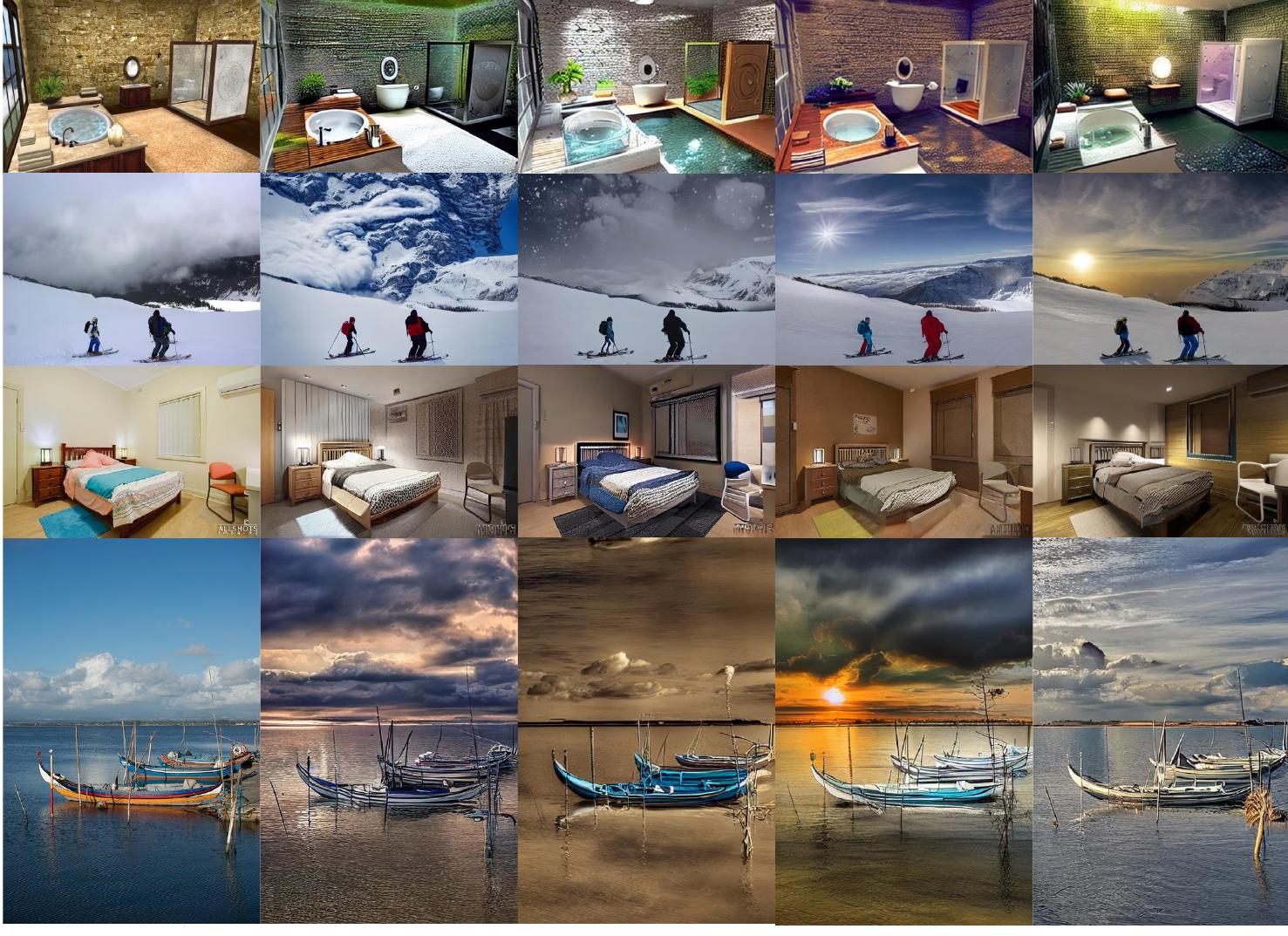}
    \caption*{(b). Synthesized data on COCO.}
    \end{minipage}
    \caption{Diffusion Synthesis can generate high-quality and diverse image-mask pairs by modifying semantics of real image (the firt column).}
    \label{fig:gen_pascal}
\end{figure*}

We first demonstrate the qualitative results of Diffusion Synthesis with masks in Fig. \ref{fig:img-mask} and in generating various scenes with different complexities in Fig. \ref{fig:gen_pascal}. Our method successfully generates diverse, realistic images while preserving the semantic information and producing sufficient details. It has the capability to manipulate high-level semantic features, such as varying weather conditions, lighting at different times of day, textures, materials, and colors.


\subsection{Discussion: Why not SAM directly?}
While the impressive performance of SAM might lead some to question the necessity of a semantic dataset synthesis pipeline, we argue that such a pipeline remains valuable in specific application scenarios. In many narrow-domain or emerging-domain use cases, where the open-vocabulary setup of models like SAM may be unnecessary or even prone to errors, a controllable dataset synthesis approach offers clear advantages. Our method offer the potential for targeted generation of specific classes, which is particularly useful in these constrained environments. Furthermore, SAM’s significant VRAM usage and slower inference speed limit its applicability on edge devices, making lightweight and efficient expert models, more desirable for real-world deployment.


\begin{table*}[t]
\centering
    \setlength{\tabcolsep}{1mm}
    \begin{tabular}{l|lllll}
    \Xhline{2\arrayrulewidth}
    \multicolumn{1}{l|}{} & \multicolumn{5}{c}{1-shot (mIoU)} \\
    \textbf{Method} & Fold-0 & Fold-1 & Fold-2 & Fold-3 & Mean$\uparrow$ 
    \\ \hline
    HDMNet\textsuperscript{$\dagger$} \cite{peng2023hierarchical} & 42.0 & 52.2 & 49.4 & 47.4 & 47.8 
    \\
    \rowcolor{lightgray} 
    Ours (+ HDMNet) & 42.2(+0.2) & 54.2(+2.0) & 50.9(+1.5) & 48.7(+1.3) & 49.0(+1.2) 
    \\
    \Xhline{2\arrayrulewidth}
\end{tabular}
    \caption{Results of 1-shot segmentation learning on COCO-$20^i$ using mIoU (\%) evaluation metric before and after applying our method on HDMNet. $\dagger$: Reproduced following the official configuration with 10,000 episodes.}
    \label{table:fsscoco_apendix}
\end{table*}

\end{document}